\documentclass[journal]{IEEEtran}
\ifCLASSINFOpdf
\else
\fi
\IEEEoverridecommandlockouts                              

\usepackage{amsthm}
\usepackage{graphicx} 
\usepackage{epsfig} 
\usepackage{mathptmx} 
\usepackage{newtxtext,newtxmath}
\usepackage{times} 
\usepackage{amsmath} 

\usepackage{lipsum}
\usepackage{mathtools}
\usepackage{cuted}
\usepackage{arydshln}
\usepackage{subcaption}
\usepackage{multirow}
\usepackage[symbol]{footmisc}

\begin{document}
%
\title{Incremental Nonlinear Fault-Tolerant Control of\\a Quadrotor with Complete Loss of\\Two Opposing Rotors
}


\author{Sihao Sun, Xuerui Wang, Qiping Chu, and Coen de Visser 
\thanks{The authors are with Control and Simulation Section, Faculty of Aerospace Engineering, Delft University of Technology, 2629 HS Delft, The Netherlands (e-mail: s.sun-4@tudelft.nl; x.wang-6@tudelft.nl; q.p.chu@tudelft.nl; c.c.devisser@tudelft.nl)}} %


%
%

\markboth{IEEE Transactions on Robotics. Preprint Version. DOI: 10.1109/TRO.2020.3010626. \copyright~2020 IEEE}%
{Shell \MakeLowercase{\textit{et al.}}: Bare Demo of IEEEtran.cls for IEEE Journals}
%




\maketitle


\begin{abstract}
In order to further expand the flight envelope of quadrotors under actuator failures, we design a nonlinear sensor-based fault-tolerant controller to stabilize a quadrotor with failure of two opposing rotors in the high-speed flight condition (> 8m/s). The incremental nonlinear dynamic inversion (INDI) approach which excels in handling model uncertainties is adopted to compensate for the significant unknown aerodynamic effects. The internal dynamics of such an underactuated system have been analyzed, and subsequently stabilized by re-defining the control output. The proposed method can be generalized to control a quadrotor under single-rotor-failure and nominal conditions. For validation, flight tests have been carried out in a large-scale open jet wind tunnel. The position of a damaged quadrotor can be controlled in the presence of significant wind disturbances. A linear quadratic regulator (LQR) approach from the literature has been compared to demonstrate the advantages of the proposed nonlinear method in the windy and high-speed flight condition.
\end{abstract}


%
\IEEEpeerreviewmaketitle

\section{INTRODUCTION}

Multi-rotor drones have demonstrated their ability in a large variety of applications such as surveillance, delivery, and recreation. Due to the potential growth of the drone market in the coming decades, safety issues are of critical concern. Apart from sensor redundancies, and improving operational regulations, fault-tolerant control (FTC) is a key to improving safety in the face of unexpected structural and actuator failures.

\begin{figure} [t!]
    \centering
    \includegraphics[scale = 0.05]{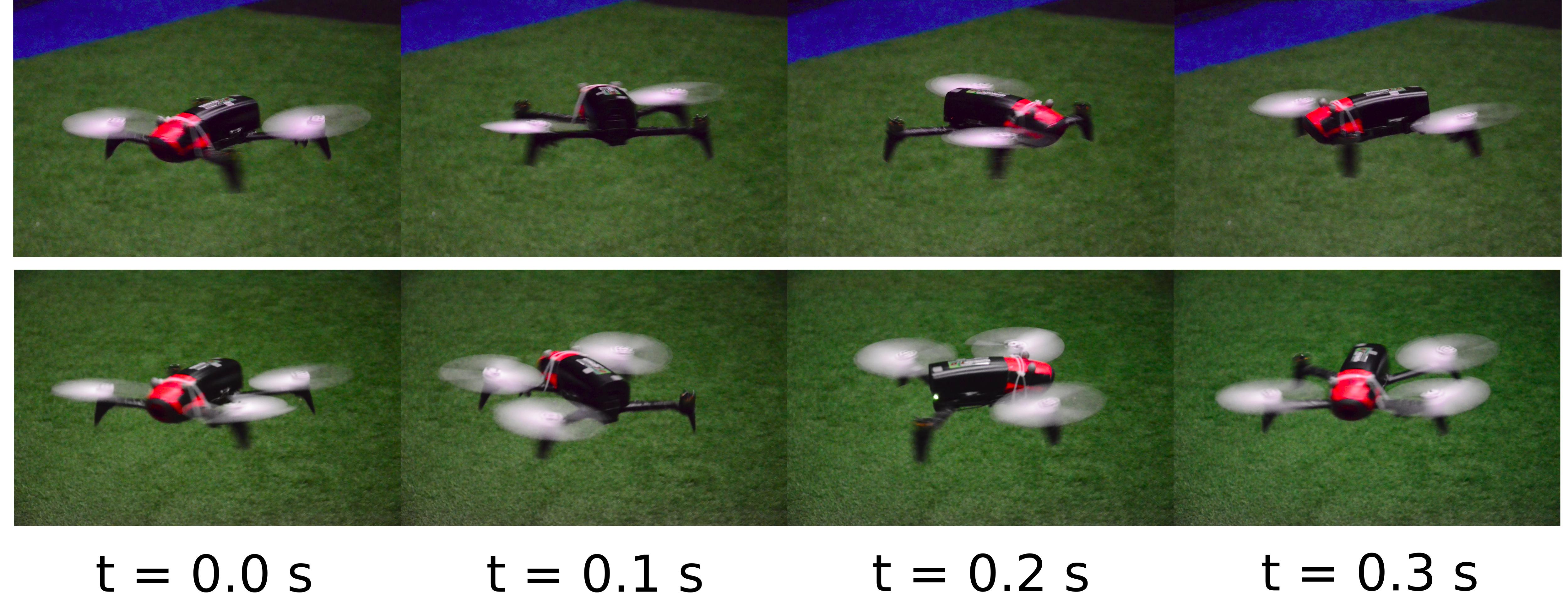}
    \caption{Snapshot of the tested Bebop2 subjected to failures of rotor 1 \& 3 (upper row), and subjected to failure of rotor 3 (bottom row).}
    \label{fig:snapshot}
\end{figure}

Among different types of multi-rotor drones, quadrotors excel in their structural simplicity. However, they suffer more from actuator damages due to a lack of actuator redundancy. Partial damage on the rotors could result in the reduction of control effectiveness, which has been extensively studied in the literature (e.g., \cite{Li2012PassiveTechnique,Besnard2012QuadrotorObserver,Wang2019QuadrotorObservers,Wang2019nonsingular}). A more challenging problem is the complete loss of one or more rotors (Fig.~\ref{fig:snapshot}). Various control methodologies addressing this problem have been proposed and validated in simulations~(e.g.,~\cite{Lanzon2014FlightFailure,Lippiello2014EmergencyApproach,Lippiello2014EmergencyApproachb,Lu2015,Morozov2018EmergencyPropellers,Crousaz2015}). 

In-flight validations have been achieved by several pieces of research where linear control methods were mostly adopted, such as linear quadratic regulator (LQR)~\cite{Mueller2014}, proportional-integral-derivative (PID) control~\cite{Merheb2017EmergencyRotor} and linear parameter varying (LPV) control~\cite{Stephan2018LinearLoss}. The relaxed hovering solution proposed by \cite{Mueller2015} indicates that the hovering flight of a quadrotor is possible with a loss of up to three rotors. With a specially designed configuration, a vehicle with only a single rotor is tested using LQR with actuator saturations taken into account~\cite{Zhang2016} .

\begin{figure}
    \centering
    \includegraphics[scale = 0.18]{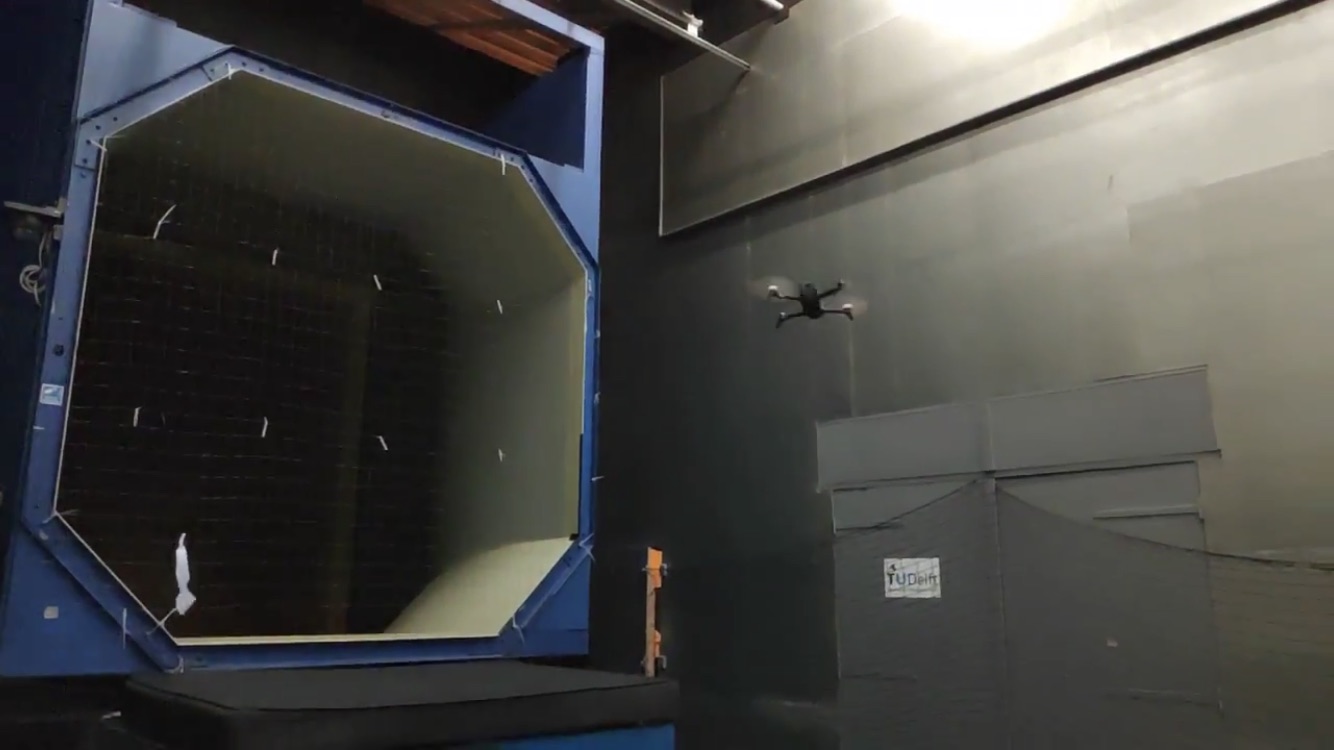}
    \caption{Snapshot of the tested quadrotor in the wind tunnel, with removal of rotor 1 and 3. The flight video can be found via \textbf{https://youtu.be/-4rXX4D5HlA}}
    \label{fig:drone_OJF}
\end{figure}

The aforementioned literature assumes that the drone is operated around the hovering condition and only limited aerodynamic effects are considered such as the rotational damping~\cite{Stephan2018LinearLoss,Mueller2015}. However, in out-door applications, significant aerodynamic forces/moments on the quadrotor are present due to fast cruising speed and large wind disturbances~\cite{Russell2016,Sun2018b}. The system nonlinearity also becomes more significant due to the complex variation of rotor aerodynamic characteristics in high-speed conditions. Therefore, designing a high-speed capable robust nonlinear controller is essential for expanding the flight envelope of a quadrotor subjected to rotor failures, and increasing its robustness against wind disturbances.

Incremental nonlinear dynamic inversion (INDI) is a sensor-based nonlinear control approach that makes use of sensor measurements to reduce its model dependency, thereby improving its robustness against model uncertainties. This approach has been adopted by the aviation industry in several applications, such as the control of fixed-wing aircraft \cite{Sieberling2010RobustPrediction}, spacecrafts~\cite{Acquatella2012RobustInversion.}, helicopters~\cite{Simplicio2013AnInversion} and multi-rotor drones~\cite{Smeur2016adaptive, Smeur2016b, Tal2019}. In \cite{Sun2018}, we made use of the INDI controller to control a quadrotor with a single rotor failure in the wind tunnel. The control method has shown its advantage in providing robustness to large aerodynamic disturbances while simplifying gain tuning, and eliminating the need to calculate an equilibrium for linear controller design.

However, the INDI controller relies on a dynamic inversion step. This step requires the number of inputs to be no less than the number of outputs. For a quadrotor with only two opposing rotors remain, the attitude control problem becomes under-actuated where the direct inversion is inapplicable. For this reason, we need to redesign the original control outputs of a quadrotor such as the thrust and attitudes. This subsequently results in several internal dynamics of which the stability needs to be guaranteed. The selection of the outputs ensuring stable internal dynamics has been addressed on some under-actuated control problems, e.g., wheeled mobile robots \cite{Yun1993InternalRobot}, quadrotor position control \cite{Lewis2009DynamicControl}, and the attitude control of space aircraft \cite{Wallner2003AttitudeDynamics}.

The main theoretical contributions of this research are twofold: (1) A detailed analysis of the internal dynamics of quadrotors with complete loss of two opposing rotors; (2) A subsequent novel robust fault-tolerant control method implementing the INDI approach. The controlled quadrotor thereby suffers less from model uncertainties caused by significant aerodynamic effects during high-speed flight. 

To validate the proposed controller, flight tests of a quadrotor with failure of two opposing rotors have been performed in an open jet wind tunnel (Fig.~\ref{fig:drone_OJF}). With limited information on the model, the controller is able to stabilize the damaged quadrotor in wind of over 8~m/s, which is more than half of its nominal maximum flight speed. This could significantly increase the safety of quadrotors by expanding the flight envelope under actuator failure conditions. With slight adaptation, the same control scheme can be used on a quadrotor with a single rotor failure for which the internal dynamics are proved to be inherently stable. A benchmark approach (LQR) has been compared to demonstrate the advantage of the proposed controller in the high-speed and windy flight conditions.

This paper is organized as follows. Section~\ref{sec:problem_formulation} provides information on the quadrotor model and the reduced attitude control. Section~\ref{sec:Incremental_nonlinear_dynamic_inversion} introduces the INDI controller and Section~\ref{sec:controller_design} directly provides the detailed controller design for a quadrotor with failure of two opposing rotors. Section~\ref{sec:stability_analysis} elaborates on the selection of control outputs and the stability of internal dynamics. Section~\ref{sec:towradsAUnifiedApproach} generalizes the proposed method to the single-rotor-failure and the nominal conditions. Finally, Sections~\ref{sec:experiment} and \ref{sec:windtunnel} demonstrate the flight test results in low-speed and high-speed flight conditions respectively.

\section{Problem Formulation}
\label{sec:problem_formulation}
\subsection{Quadrotor Kinematic and Dynamic Model}
There are two coordinate systems considered in this work. The inertial frame $\mathcal{F_I}=\{O_I,\boldsymbol{x}_I,\boldsymbol{y}_I,\boldsymbol{z}_I\}$, is fixed to the ground, with $\boldsymbol{x}_I$, $\boldsymbol{y}_I$ and $\boldsymbol{z}_I$ pointing to the north, east and aligning with the local gravity. The body frame $\mathcal{F_B}=\{O_B,\boldsymbol{x}_B,\boldsymbol{y}_B,\boldsymbol{z}_B\}$ is fixed to the vehicle, with the origin located at the center of mass. 
As Fig.~1 shows, we assume the quadrotor has a symmetric fuselage, which is a common configuration for many commercially available quadrotors. As a convention, we define $\boldsymbol{x}_B$ points forward, $\boldsymbol{z}_B$ points downwards such that the drone inertia is symmetric with respect to the $\boldsymbol{x}_B-\boldsymbol{z}_B$ plane, and $\boldsymbol{z}_B$ is parallel with the thrust direction. $\boldsymbol{y}_B$ thus points rightwards to render $\mathcal{F_B}$ a right-handed coordinate system.
In the following context, the superscript $[\cdot]^I$ and $[\cdot]^B$ indicate the coordinate system in which a vector is expressed.

The equations of motion of a quadrotor are formulated as follows:
\begin{equation}
\dot{\boldsymbol{P}}^I = \boldsymbol{V}^I
\label{eq:P_dot}
\end{equation}
\begin{equation}
m_v\dot{\boldsymbol{V}}^I = m_v\textbf{\textit{g}}^I+\boldsymbol{RF}^B
\label{eq:V_dot}
\end{equation}
\begin{equation}
\dot{\boldsymbol{R}}=\boldsymbol{R}\boldsymbol{\Omega}^B_{\times}
\end{equation}
\begin{equation}
\boldsymbol{I_v} \dot{\boldsymbol{\Omega}}^B = -\boldsymbol{{\Omega}}^B_{\times} \boldsymbol{I_v\Omega}^B + \boldsymbol{M}^B
\label{eq:pqr_dot}
\end{equation}
where $\boldsymbol{P}^I=[X,~Y,~Z]^T$ and $\boldsymbol{V}^I=[V_x,~V_y,~V_z]^T$ represent the position and the velocity of the center of mass in $\mathcal{F_I}$; $m_v$ is the vehicle gross mass and $\boldsymbol{I_v}$ denotes the inertia matrix of the vehicle including rotors. $\boldsymbol{g}$ is the local gravity vector. $\boldsymbol{R}\in \mathrm{SO}(3)$ indicates the transformation matrix from $\mathcal{F_B}$ to $\mathcal{F_I}$. The angular velocity is expressed as $\boldsymbol{\Omega}^B = [p~q~r]^T$ where $p$, $q$ and $r$ denote pitch rate, roll rate and yaw rate respectively. $\boldsymbol{\Omega}_{\times}$ is the skew symmetric matrix such that $\boldsymbol{\Omega}_{\times}\boldsymbol{a} =  \boldsymbol{\Omega}\times \boldsymbol{a}$ for any vector $\boldsymbol{a}\in \mathbb{R}^3$. 

\begin{figure}
\centering
\includegraphics[scale = 0.5]{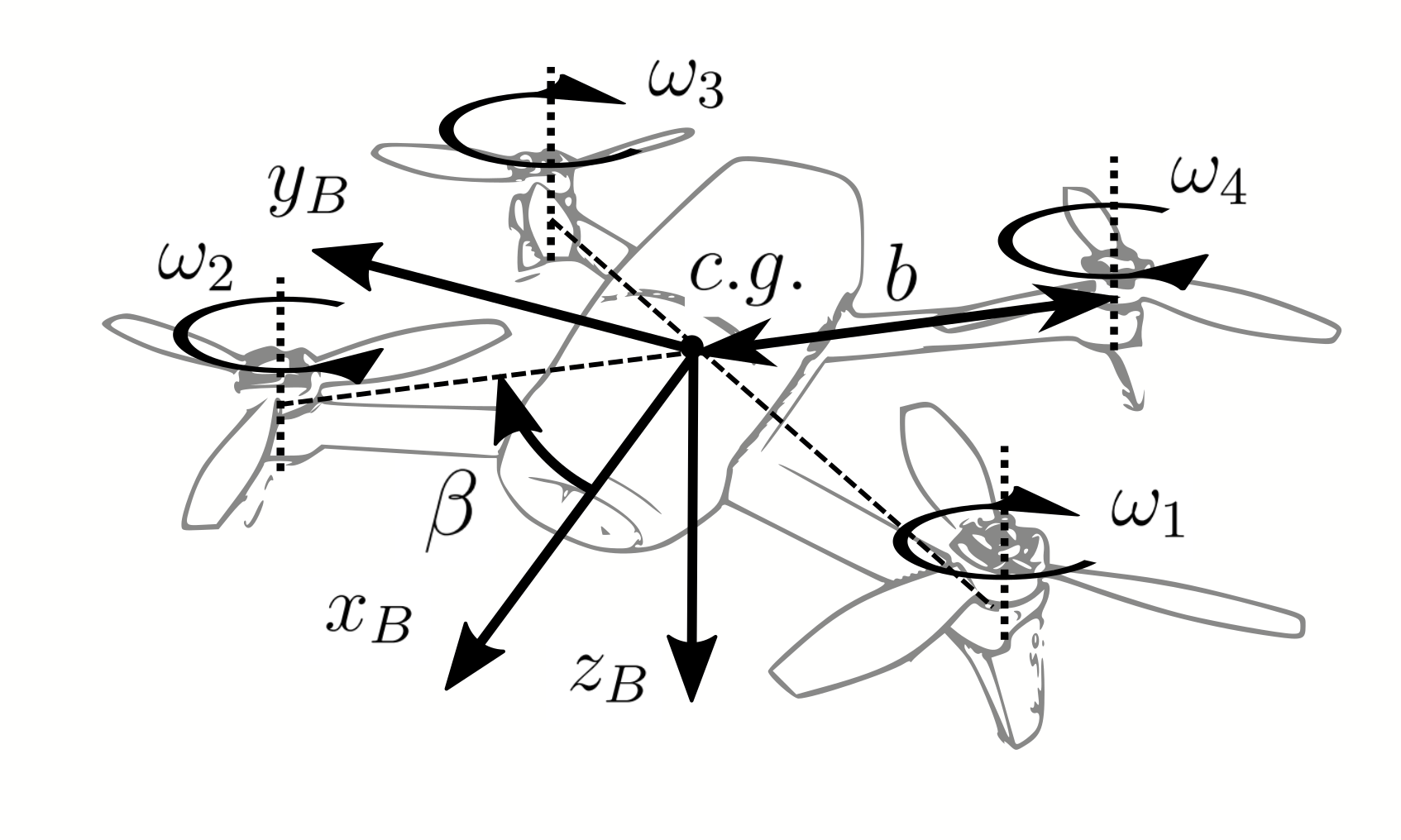}
\caption{Definition of geometry parameters, rotor index and directions, and the body frame $\mathcal{F_B}$.}
\label{fig:drone_drawing}
\end{figure}

The variables $\boldsymbol{F}^B$ and $\boldsymbol{M}^B$ denote the resultant force and moment on the center of mass respectively, projected on $\mathcal{F_B}$. For a quadrotor with thrust parallel to the $\boldsymbol{z}_B$ axis and rotor directions shown in Fig.~\ref{fig:drone_drawing}, we have

\begin{equation}
    \boldsymbol{F}^B = \left[\begin{array}{c}
         0  \\
         0  \\
         -\bar{\kappa}\sum_{i=1}^4\omega_i^2
    \end{array}\right] + \boldsymbol{F}_a
    \label{eq:F}
\end{equation}
\begin{equation}
\resizebox{1.05\hsize}{!}{$
\begin{array}{rl}
     \boldsymbol{M}^B &= 
     \bar{\kappa}\left[\begin{array}{c c c c}
b\sin\beta & -b\sin\beta & -b\sin\beta & b\sin\beta\\
b\cos\beta & b\cos\beta & -b\cos\beta & -b\cos\beta\\
\sigma & -\sigma & \sigma & -\sigma
\end{array}\right]
\left[\begin{array}{c}
     \omega_1^2  \\
     \omega_2^2  \\
     \omega_3^2  \\
     \omega_4^2
\end{array}\right]  \\
     &+  \left[\begin{array}{c}
         I_pq(\omega_1-\omega_2+\omega_3-\omega_4)  \\
         -I_pp(\omega_1-\omega_2+\omega_3-\omega_4)  \\
         I_p(\dot{\omega}_1-\dot{\omega}_2+\dot{\omega}_3-\dot{\omega}_4)
    \end{array}\right] + 
    \left[\begin{array}{c}
        0\\
        0\\
        -\gamma r
    \end{array}\right] +\boldsymbol{M}_a
\end{array}$}
\label{eq:M}
\end{equation}
where $\bar{\kappa}$ is a thrust coefficient valid in the hovering condition; $\sigma$ is a constant ratio between the thrust coefficient and drag coefficient of the rotor; $b$ and $\beta$ are geometry parameters as Fig.~\ref{fig:drone_drawing} shows. Note that $\beta\in(0,~\pi/2)$ for a quadrotor. $\boldsymbol{\omega}^B_{i}=[0,~0,~\omega_{i}]$ is the angular speed of the $i$-th rotor with respect to the body. $I_p$ denotes the moment of inertia of each rotor about the rotational axis.  Note that this model assumes that $||\boldsymbol{\Omega}||<<||\boldsymbol{\omega}_i||$, thus the magnitude of rotor angular speed with respect to the air is approximated by $\omega_i$. Symbol $||\cdot||$ is defined as the $L^2$ norm of a vector. $\gamma$ in (\ref{eq:M}) indicates the aerodynamic yaw damping coefficient~\cite{Mueller2014,Stephan2018LinearLoss}. 

In the high-speed flight condition, there are significant aerodynamic effects such as thrust variation \cite{Hoffmann2011}, rotor in-plane force \cite{Mahony2012}, rotor moment and airframe aerodynamic \cite{Sun2019Aero}. These additional aerodynamic related forces and moments are then expressed as $\boldsymbol{F}_a$ and $\boldsymbol{M}_a$ in (\ref{eq:F}) and (\ref{eq:M}). They are regarded as model uncertainties that need to be compensated for by the robustness of the control method.

\subsection{Reduced Attitude Control}
\label{sec:reducedattitudecontrol}
The concept of reduced attitude control \cite{Fortescue2011Rigid-BodyControl} has been adopted by \cite{Mueller2015} in the quadrotor fault tolerant control problem. We hereby briefly introduce the concept.

For a quadrotor with complete rotor failures, the full state equilibrium becomes unattainable. This is due to the incapacity of the remaining rotors to generate zero yaw moment while producing necessary thrust. As a consequence, the vehicle spins around the yaw axis. And the attitude control is reduced to a thrust vector pointing problem without considering the yaw angle.

Define a unit vector $\boldsymbol{n}$ fixed to $\mathcal{F_B}$ where $\boldsymbol{n}^B=[n^B_x,~n^B_y,~n^B_z]^T$. For a quadrotor with failure of two opposing rotors, choosing $\boldsymbol{n}^B=[0,~0,~-1]^T$ is most energy-efficient \cite{Mueller2014} where $\boldsymbol{n}$ aligns with the instantaneous thrust direction. 
Define another unit vector $\boldsymbol{n}_d$ as the reference of $\boldsymbol{n}$, which is calculated by the position controller or remotely provide by a pilot. Then aligning $\boldsymbol{n}_d$ with $\boldsymbol{n}$ (or vise versa) becomes the primary task of the attitude controller. Therefore, we introduce the following relaxed attitude kinematic equation:~\cite{Sun2018}
\begin{equation}
\dot{\boldsymbol{n}}_d^B = -\boldsymbol{\Omega_{\times}}\boldsymbol{n}_d^B + \boldsymbol{R}^T\dot{\boldsymbol{n}}_d^I
\label{eq:relaxed_att_kine}
\end{equation}
With the expressions $\boldsymbol{n}_d^B=[h_1,~h_2,~h_3]^T$ and $ \boldsymbol{R}^T\dot{\boldsymbol{n}}_d^I = [\lambda_1,~\lambda_2,~\lambda_3]^T$, the expanded formula of (\ref{eq:relaxed_att_kine}) can be given as:
\begin{equation}
    \left[\begin{array}{c}
         \dot{h}_1  \\
         \dot{h}_2 \\
         \dot{h}_3
    \end{array}\right]
=\left[
\begin{array}{ccc}
     0 & r & -q  \\
     -r &  0 & p \\
     q & -p & 0 
\end{array}\right]
\left[
\begin{array}{c}
     h_1  \\
     h_2  \\
     h_3  \\
\end{array}
\right] + 
\left[
\begin{array}{c}
     \lambda_1  \\
     \lambda_2  \\
     \lambda_3  \\
\end{array}
\right]
\label{eq:relaxed_att_kine_expand}
\end{equation}

To align $\boldsymbol{n}$ with $\boldsymbol{n}_d$, we can control $h_1$ and $h_2$ to track $n_x^B$ and $n_y^B$ respectively. To be specific, with the selection of $\boldsymbol{n}^B=[0,~0,~-1]^T$, $h_1$ and $h_2$ need to be stabilized to zero. The other selections of $\boldsymbol{n}^B$ may be considered for the case with single rotor failure, which has been discussed in \cite{Mueller2014,Mueller2015}.  

The challenge of the problem is conducting relaxed attitude control of a quadrotor with only two opposing rotors remain in the presence of significant model uncertainties $\boldsymbol{F}_a$ and $\boldsymbol{M}_a$, for instance, in high-speed flight conditions where significant aerodynamic effects become apparent. To achieve this goal, we employ a sensor-based nonlinear control method to be described in Sec.~\ref{sec:Incremental_nonlinear_dynamic_inversion}. The detailed implementation of this method will be provided in Sec.~\ref{sec:controller_design}.

\section{Methodology}
\label{sec:Incremental_nonlinear_dynamic_inversion}
Incremental nonlinear dynamic inversion (INDI) is a sensor-based nonlinear control approach. The approach stems from the nonlinear dynamic inversion (NDI) control. INDI reduces the model dependencies of NDI by replacing non-input related model terms with direct sensor measurements, or sensor measurement derived quantities, thereby greatly improving robustness against model uncertainties.

For aircraft systems, including nominal quadrotors, each sub-problem (e.g., the attitude and rate control loops) is fully actuated without internal dynamics to be analyzed~\cite{Wang2019QuadrotorObservers,Sieberling2010RobustPrediction, Smeur2016adaptive}. However, for a quadrotor with failure of two opposing rotors, the number of control inputs is less than the required output in a conventional cascaded control setup, yielding internal dynamics that have to be stabilized. 
In the following context, the generalized INDI control considering internal dynamics will be briefly reviewed. Readers may refer to \cite{Wang2018StabilityControl} and \cite{Khalil2002NonlinearSystems} for further details.

Consider a nonlinear input-affine system
\begin{equation}
\begin{split}
    \boldsymbol{\dot{x}} &=\boldsymbol{f}(\boldsymbol{x})+\boldsymbol{G}(\boldsymbol{x})\boldsymbol{u}\\
    \boldsymbol{y} &= \boldsymbol{h}(\boldsymbol{x})
    \label{eq:nonlinear_problem}
\end{split}
\end{equation}
where $\boldsymbol{f}: \mathbb{R}^{n}\rightarrow \mathbb{R}^n$ and $\boldsymbol{h}:\mathbb{R}^n\rightarrow \mathbb{R}^l$ are smooth vector fields. $\boldsymbol{G}: \mathbb{R}^{n}\rightarrow \mathbb{R}^{n\times m}$ is a function mapping with smooth vector fields as columns. The number of outputs is not larger than the number of inputs (i.e., $l\leq m$). There exists a nonlinear transformation $\boldsymbol{T}: \mathbb{R}^{n}\rightarrow \mathbb{R}^n$ such that the states $\boldsymbol{x}$ can be transformed to the normal form including internal states $\boldsymbol{\eta}$ and external states $\boldsymbol{\xi}$:
\begin{equation}
\left[
\begin{array}{c}
     \boldsymbol{\eta}\\
     \boldsymbol{\xi} 
\end{array}
\right]
=
\left[
\begin{array}{c}
     \boldsymbol{\phi}(\boldsymbol{x})\\
     \boldsymbol{\theta}(\boldsymbol{x}) 
\end{array}
\right]
=
\boldsymbol{T}(\boldsymbol{x})
\label{eq:transformation}
\end{equation}
where
\begin{equation}
\boldsymbol{\theta}(\boldsymbol{x}) = [\boldsymbol{\theta}_1(\boldsymbol{x}),~\boldsymbol{\theta}_2(\boldsymbol{x}),...,~\boldsymbol{\theta}_l(\boldsymbol{x})]^T
\label{eq:external_states_transformation1}
\end{equation}
with
\begin{equation}
    \boldsymbol{\theta}_i(\boldsymbol{x}) = [h_i(\boldsymbol{x}),~L_fh_i(\boldsymbol{x}),...,L_f^{\rho_i-1}h_i(\boldsymbol{x})],~i=1,2,...,l
    \label{eq:external_states_transformation2}
\end{equation}
where $h_i(\boldsymbol{x})$ indicates the $i$th element in the vector field $\boldsymbol{h}$. The notation $L_f^{\rho_i}h_i(\boldsymbol{x})$ indicates the $\rho_i$th order Lie derivative of the function $h_i$ with respect to the vector fields $\boldsymbol{f}(\boldsymbol{x})$~\cite{Khalil2002NonlinearSystems}. $\rho_i$ indicates the relative degree of the $i$th output $y_i$.

By defining $\bar{\rho}=\Sigma_1^l\rho_i$ as the sum of relative degrees of each output, one can define the transformation $\boldsymbol{\phi}(\boldsymbol{x}) = [\phi_1(\boldsymbol{x}),~\phi_2(\boldsymbol{x}),~...,~\phi_{n-\bar\rho}(\boldsymbol{x})]$. The selection of $\boldsymbol{\phi}(\boldsymbol{x})$ is not unique, but has to satisfy the following condition:
\begin{equation}
    \frac{\partial\phi_i}{\partial\boldsymbol{x}}\boldsymbol{G}(\boldsymbol{x}) = 0,\qquad i=1,2,...,n-\bar\rho
    \label{eq:phi_condition}
\end{equation}
namely the first-order derivative of $\boldsymbol{\eta}$ as defined per (\ref{eq:transformation}) does not include control input $\boldsymbol{u}$. The nonlinear transformation $\boldsymbol{T}(\boldsymbol{x})$ is a diffeomorphism (i.e., smooth and invertible) in the domain of interest. 

As a consequence, the problem is transformed to the normal form
\begin{equation}
\begin{split}
    \dot{\boldsymbol{\eta}} &= \boldsymbol{f}_{\eta}(\boldsymbol{\eta},\boldsymbol{\xi})\\
    \dot{\boldsymbol{\xi}} &= \boldsymbol{A_c}\boldsymbol{\xi} + \boldsymbol{B_c}[\boldsymbol{\alpha}(\boldsymbol{x})+\boldsymbol{\mathcal{B}}(\boldsymbol{x})\boldsymbol{u}]\\
    \boldsymbol{y} &= \boldsymbol{C_c}\boldsymbol{\xi}
\end{split}
\label{eq:normal_form}
\end{equation}
where the triplet $(\boldsymbol{A_c},\boldsymbol{B_c},\boldsymbol{C_c})$ is a canonical form representation of $l$ chains of $\rho_i$ integrators ($i=1,2,...,l$); $\boldsymbol{\alpha}: \mathbb{R}^n\rightarrow \mathbb{R}^l$ and $\boldsymbol{\mathcal{B}}:\mathbb{R}^n\rightarrow \mathbb{R}^{l\times m}$ are mappings determined by the system (\ref{eq:nonlinear_problem}). Subsequently, the output dynamics can be represented as
\begin{equation}
    \boldsymbol{y}^{(\boldsymbol{\rho})} = \boldsymbol{\alpha}(\boldsymbol{x})+\boldsymbol{\mathcal{B}}(\boldsymbol{x})\boldsymbol{u}
    \label{eq:output_dynamics}
\end{equation}
where $\boldsymbol{y}^{(\boldsymbol{\rho})}=[y_1^{(\rho_1)},~y_2^{(\rho_2)},...,~y_l^{(\rho_l)}]^T$. 
If matrix $\boldsymbol{\mathcal{B}}$ has full row rank, the NDI control law can be designed as 
\begin{equation}
    \boldsymbol{u}_\mathrm{ndi} = \boldsymbol{\mathcal{B}}(\boldsymbol{x})^+(\boldsymbol{\nu}-\boldsymbol{\alpha}(\boldsymbol{x}))
    \label{eq:NDI_control_law}
\end{equation}
where superscript $[\cdot]^+$ indicates the Moore-Penrose inverse of the matrix; $\boldsymbol{\nu} \in \mathbb{R}^n$ is called the pseudo-input. With a full knowledge of $\boldsymbol{\alpha}(x)$ and $\boldsymbol{\mathcal{B}(x)}$, control law (\ref{eq:NDI_control_law}) yields the closed loop dynamics
\begin{equation}
    \boldsymbol{y}^{(\boldsymbol{\rho})} = \boldsymbol{\nu}
\end{equation}
For a command tracking problem with reference output $\boldsymbol{y}_\mathrm{ref}\in \mathbb{R}^l$ that is $\boldsymbol{\rho}$th order differentiable, selecting
\begin{equation}
\boldsymbol{\nu}=-\boldsymbol{K}(\boldsymbol{\xi}-\boldsymbol{\xi}_\mathrm{ref}) + \boldsymbol{y}_\mathrm{ref}^{(\boldsymbol{\rho})}
\label{eq:nu_ndi}
\end{equation}
ensures that the reference output is being tracked asymptotically, where the gains $\boldsymbol{K}$ is selected such that $\boldsymbol{A_c}-\boldsymbol{B_c}\boldsymbol{K}$ is Hurwitz. The reference $\boldsymbol{\xi}_\mathrm{ref}$ is denoted as
\begin{equation}
\begin{split}
        \boldsymbol{\xi}_\mathrm{ref} &= [\boldsymbol{\psi}_1,~\boldsymbol{\psi}_2,...,\boldsymbol{\psi}_l]^T,\\
        \boldsymbol{\psi}_i &= [y_{\mathrm{ref},i},~y_{\mathrm{ref},i}^{(1)},...,y_{\mathrm{ref},i}^{(\rho_i-1)}],~i = 1,2,...,l
\end{split}
\end{equation}

In reality, the nonlinear model dependent terms $\boldsymbol{\alpha}(\boldsymbol{x})$ and $\boldsymbol{\mathcal{B}}(\boldsymbol{x})$ are almost impossible to be obtained due to inevitable model uncertainties. In view of this, we take the first-order Taylor series expansion of (\ref{eq:output_dynamics}) around the condition at the last sensor sampling moment $t-\Delta t$ (denoted by subscript $[\cdot]_0$), then (\ref{eq:output_dynamics}) becomes
\begin{eqnarray}
\boldsymbol y^{(\boldsymbol \rho)} &=& \boldsymbol \alpha (\boldsymbol x) + \boldsymbol{\mathcal{B}}(\boldsymbol x) \boldsymbol u  \nonumber \\
&=& \boldsymbol y^{(\boldsymbol \rho)}_0 + \boldsymbol{\mathcal{B}}(\boldsymbol x_0)\Delta \boldsymbol u  +\frac{\partial [\boldsymbol \alpha(\boldsymbol x) + \boldsymbol{\mathcal{B}}(\boldsymbol x)\boldsymbol u]}{\partial \boldsymbol x}\bigg|_0 \Delta \boldsymbol x + \boldsymbol{R}_1 \qquad
\label{eq:INDI_formula}
\end{eqnarray}
where $\Delta\boldsymbol{u} = \boldsymbol{u}-\boldsymbol{u}_0,~\Delta\boldsymbol{x} = \boldsymbol{x}-\boldsymbol{x}_0$; $\boldsymbol{R}_1$ is the first-order Taylor expansion remainder. Design the incremental nonlinear dynamic inversion (INDI) control as 
\begin{equation}
    \bar{\boldsymbol{u}}_\mathrm{indi}=\boldsymbol{\mathcal{\hat{B}}}(\boldsymbol{x}_0)^+(\boldsymbol{\nu}-\boldsymbol{y}_0^{(\boldsymbol{\rho})}) + \boldsymbol{u}_0
    \label{eq:control_INDI0}
\end{equation}
where $\boldsymbol{\nu}$ is selected as per (\ref{eq:nu_ndi}), while $\boldsymbol{\mathcal{\hat{B}}}$ is the estimated control effectiveness matrix. As a result, the closed-loop tracking error ($\boldsymbol{e} = \boldsymbol{y} - \boldsymbol{y}_{\text{ref}}$) dynamics are $\dot{\boldsymbol e} 
= (\boldsymbol A_c -\boldsymbol B_c \boldsymbol K) \boldsymbol e + \boldsymbol B_c \boldsymbol \varepsilon_{\text{indi}}$, where $\boldsymbol \varepsilon_{\text{indi}}$ is the residual error caused by model uncertainties, $\Delta \boldsymbol{x}$-related term and $\boldsymbol{R}_1$ in (\ref{eq:control_INDI0}). The ultimate boundedness of $\boldsymbol \varepsilon_{\text{indi}}$ and $\boldsymbol{e}$ has been proved in~\cite{Wang2019SMC}.

In INDI control, the model information of $\boldsymbol{\alpha}(\boldsymbol{x})$ required for NDI controller, is not needed for implementation, which greatly reduces the effort of modeling. The control effectiveness matrix $\boldsymbol{\mathcal{\hat{B}}}$ is relatively easier to be estimated offline or identified online~\cite{Smeur2016adaptive}. Apart form its reduced model dependency, INDI control also has enhanced robustness as compared to its classical NDI counterpart~\cite{Wang2019SMC}.

Due to the measurement noise, the variables $\boldsymbol{x}_0$, $\boldsymbol{y}_0$ can be low-pass filtered in practice. To synchronize the time delay caused by these filters, $\boldsymbol{u}_0$ also need to be filtered with the same cut-off frequency~\cite{Smeur2016adaptive}. We use subscript $[\cdot]_f$ to denote the filtered variables (e.g., $\boldsymbol{x}_0\rightarrow\boldsymbol{x}_f$, $\boldsymbol{y}_0\rightarrow\boldsymbol{y}_f$ and $\boldsymbol{u}_0\rightarrow\boldsymbol{u}_f$). Consequently, the INDI control law becomes
\begin{equation}
    \boldsymbol{u}_\mathrm{indi}=\boldsymbol{\mathcal{\hat{B}}}(\boldsymbol{x}_f)^+(\boldsymbol{\nu}-\boldsymbol{y}_f^{(\boldsymbol{\rho})}) + \boldsymbol{u}_f
    \label{eq:control_INDI}
\end{equation}
We will elaborate on applying the INDI control law (\ref{eq:control_INDI}) to the quadrotor control problem in the following sections. 

\section{Controller Design}
\label{sec:controller_design}
\begin{figure}
    \centering
    \includegraphics[scale=0.58]{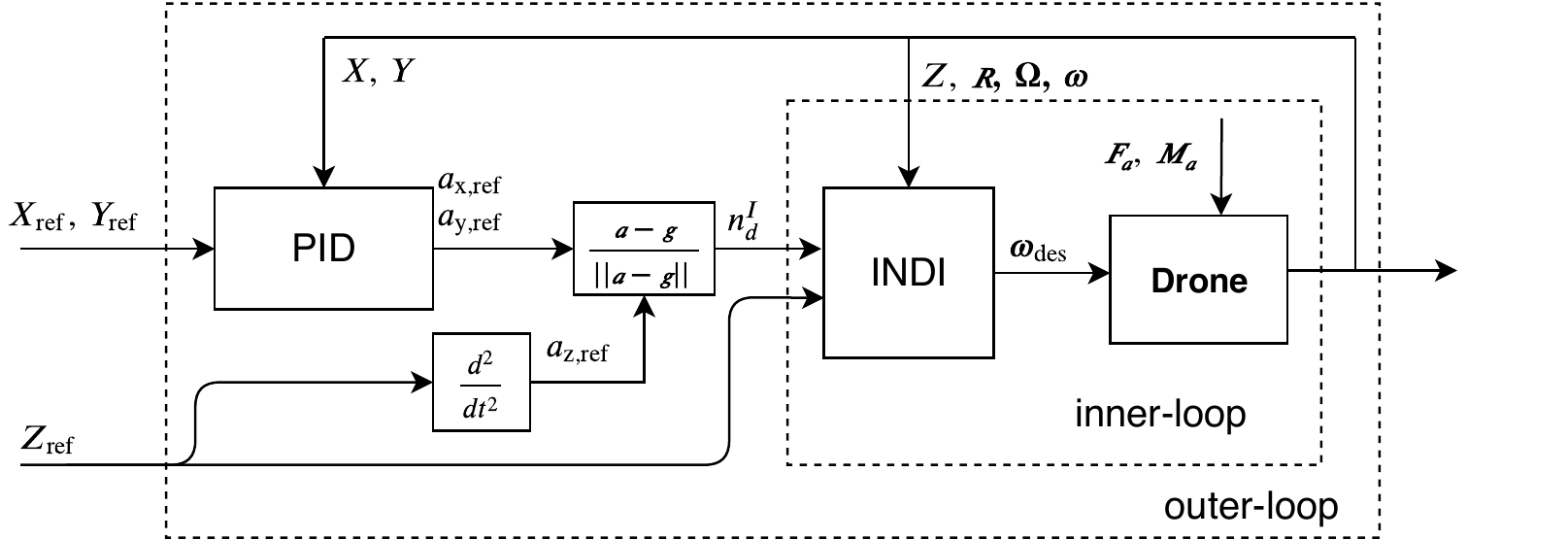}
    \caption{The two-loops cascaded control scheme using PID as outer-loop position control  and INDI as inner-loop altitude / attitude control.}
    \label{fig:control_scheme_DRF}
\end{figure}
The detailed design process of the controller for a quadrotor with complete failure of two opossing rotors is presented in this section. In general, we use a cascaded controller with two loops (Fig.~\ref{fig:control_scheme_DRF}), where INDI is applied in the inner-loop for compensating model uncertainties ($\boldsymbol{M}_a,~\boldsymbol{F}_a$).

\subsection{Outer-loop Design}
The outer-loop contains a horizontal position controller that computes the acceleration command from the reference horizontal position denoted by $X_\mathrm{ref}$ and $Y_\mathrm{ref}$. Due to the linear property of translational kinematics, a linear method such as a PID controller can be employed. In addition, the reference altitude $Z_\mathrm{ref}$ needs to be second-order differentiable. Therefore, we have
\begin{equation}
    \boldsymbol{a}_\mathrm{ref} = \left[ 
    \begin{array}{c}
    -k_pe_x - k_d\dot{e}_x - k_i\int e_xdt\\
    -k_pe_y - k_d\dot{e}_y - k_i\int e_ydt\\
    \ddot{Z}_\mathrm{ref}
    \end{array}
    \right]
    \label{eq:PID}
\end{equation}
where $e_x = X - X_\mathrm{ref}$, $e_y = Y - Y_\mathrm{ref}$ denote the horizontal position errors in $\mathcal{F_I}$; control gains $k_p$, $k_i$ and $k_d$ are positive.
Then $\boldsymbol{n}_d$ for reduced attitude control (see Sec.~\ref{sec:problem_formulation}) can be calculated by
\begin{equation}
    \boldsymbol{n}_d = \frac{\boldsymbol{a}_\mathrm{ref}-\boldsymbol{g}}{||\boldsymbol{a}_\mathrm{ref}-\boldsymbol{g}||}
\end{equation}

It is possible to replace (\ref{eq:PID}) by more sophisticated position controllers to obtain $\boldsymbol{a}_\mathrm{ref}$, which will not be elaborated in this research.
\subsection{Inner-loop Design}
The altitude control is included in the inner-loop controller since the altitude reference $Z_\mathrm{ref}$ is related to rotor thrust which contains model uncertainties $\boldsymbol{F}_a$. Consequently, the inner-loop is a combination of altitude and attitude control using the INDI approach.

States for the inner-loop control are defined as $\boldsymbol{x}_\mathrm{in} = [h_1,~h_2,~p,~q,~r,~Z,~V_z]^T$.  There are two different scenarios for a quadrotor with two opposite rotor failures. If only rotor 1 and 3 remain functional, we define the control input as
\begin{equation}
    \boldsymbol{u} = [u_1,~u_2]^T \triangleq [\omega_1^2,~\omega_3^2]^T,~~s_l=1
    \label{eq:u_define1}
\end{equation}
If only rotor 2 and 4 remain, then
\begin{equation}
    \boldsymbol{u} = [u_1,~u_2]^T \triangleq [\omega_2^2,~\omega_4^2]^T,~~s_l=-1
    \label{eq:u_define2}
\end{equation}
where $s_l\in\{-1,~1\}$ is a parameter indicating the type of failure.

Since the product of inertia is negligible compared with the moment of inertia for a typical quadrotor, we can assume a diagonal inertia matrix $\boldsymbol{I_v}=\mathrm{diag}(I_x,~I_y,~I_z)$. Thus the state equations for the inner-loop states $\boldsymbol{x}$ can be derived from (\ref{eq:pqr_dot}) and (\ref{eq:relaxed_att_kine}), yielding
\begin{equation}
\left[\begin{array}{c}
\dot{Z}\\
\dot{V}_z\\
\dot{h}_1\\
\dot{h}_2\\
\dot{p}\\
\dot{q}\\
\dot{r}
\end{array}\right]
=
\left[
\begin{array}{c}
     V_z  \\
 g + F_{a,z} - R_{33}\bar{\kappa}(u_1+u_2)/m_v\\
 h_3p-h_1r + \lambda_2\\
 -h_3q+h_2r + \lambda_1\\
 A_xrq - 2a_xq\bar{\omega}s_n + M_{a,x} + s_lG_p(u_1 - u_2)\\
 A_yrp + 2a_yp\bar{\omega}s_n + M_{a,y} + G_q(u_1-u_2)\\
 A_zpq -\gamma r/I_z + M_{a,z} - s_n G_r(u_1+u_2)
\end{array}
\right]
\label{eq:x_in_dot}
\end{equation}
where 
\begin{equation}
A_x = (I_y-I_z)/{I_x},~A_y = (I_z-I_x)/{I_y},~A_z = (I_x-I_y)/{I_z}\end{equation}
\begin{equation}a_x = I_p/I_x,~a_y = I_p/I_y,~g = ||\boldsymbol{g}||\end{equation}
where $R_{33}$ in (\ref{eq:x_in_dot}) represens the entry at the third row and the third column of the matrix $\boldsymbol{R}$; $\bar{\omega}$ is the average angular rate of the remaining rotors; $s_n\in\{-1,~1\}$ indicates the handedness of the remaining rotors with $1$ clockwise and $-1$ counterclockwise. $G_p$, $G_q$ and $G_r$ are control effectiveness on angular accelerations, where
\begin{equation}
    G_p = \bar{\kappa}\sin\beta/I_x,~~G_q = \bar{\kappa}\cos\beta/I_y,~~G_r = \sigma\bar{\kappa}/I_z
\end{equation}
Normally, $|G_r| << \min\{|G_p| ,|G_q|\}$.
Note that state equations (\ref{eq:x_in_dot}) are nonlinear and contain model uncertainties ($F_{a,z}$, $M_{a,x}$, $M_{a,y}$ and $M_{a,z}$). 
The following content in this section designs the INDI control law (\ref{eq:control_INDI}) for this specific problem.

\subsubsection{Control output definition}
Since there are only two inputs remain, we can select a maximum of two variables as control outputs. To guarantee altitude tracking, we choose the first output as
\begin{equation}
    y_1 = Z
    \label{eq:output1}
\end{equation}

\begin{figure}
    \centering
    \includegraphics[scale = 0.7]{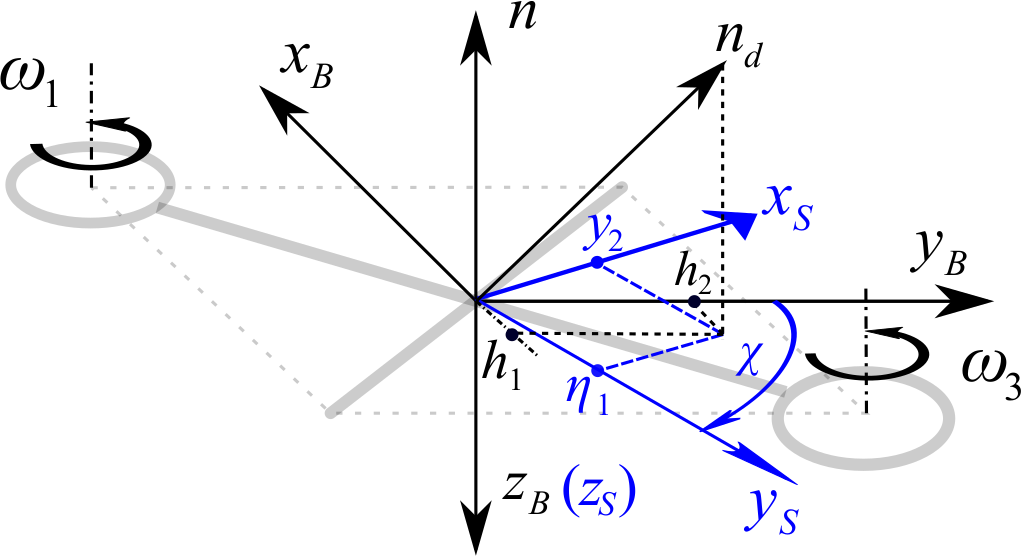}
    \caption{Definition of $\mathcal{F_S}$, $y_2$, $\eta_1$ and $\chi$ when rotor 2 and rotor 4 are removed.}
    \label{fig:def_y2_chi}
\end{figure}

The second output have to be associated with the reduced attitude control. Recall that in Sec.~\ref{sec:reducedattitudecontrol}, $\boldsymbol{n}$ needs to align with $\boldsymbol{n}_d$ by manipulating the vehicle attitude, and $h_1$ and $h_2$ of (\ref{eq:relaxed_att_kine_expand}) need to converge to zero. We hereby introduce a new coordinate system $\mathcal{F_S}=\{O_S,\boldsymbol{x}_S,\boldsymbol{y}_S,\boldsymbol{z}_S\}$ that is fixed with respect to the body frame. As Fig.~\ref{fig:def_y2_chi} illustrates, $\mathcal{F_S}$ is generated by rotating the body frame about $\boldsymbol{z}_B$. The rotation angle is denoted as $\chi$.

The second output $y_2$ is then defined as the projection of $\boldsymbol{n}_d$ on $\boldsymbol{x}_S$. In other words, $y_2$ becomes a linear combination of $h_1$ and $h_2$ scheduled by the angle $\chi$:
\begin{equation}
    y_2 = h_1\cos{\chi} + h_2\sin{\chi}
    \label{eq:output2}
\end{equation}

Meanwhile, the projection of $\boldsymbol{n}_d$ on $\boldsymbol{y}_S$ that is perpendicular to the second output $y_2$ remains uncontrolled (see Fig.~\ref{fig:def_y2_chi}). We will elaborate in Sec.~\ref{sec:stability_analysis} that this variable, denoted by $\eta_1$, is one of the internal states to be stabilized by properly selecting the angle $\chi$.

Due to the symmetric property of a quadrotor, we can determine $\chi$ by selecting its absolute value, using the following relationship:
\begin{equation}
    \chi = s_l|\chi|
\end{equation} 
It is noteworthy that $|\chi|$ is associated with the control performance, which can be analogous to physically informed control gains. The selection of $|\chi|$ will be further discussed in Sec.~\ref{sec:stability_analysis}.

\subsubsection{Control effectiveness estimation}
After defining the outputs we can take second order derivative of both $y_1$ and $y_2$, yielding
\begin{equation}
\begin{split}
    \ddot{y}_1 &= g + F_{a,z} - R_{33}\bar{\kappa}(u_1+u_2)/m_v\\ &=\alpha_1 + B_1(u_1+u_2)    
\end{split}
\label{eq:ddoty1}
\end{equation}
\begin{equation}
\begin{split}
    \ddot{y}_2 & = \ddot{h}_1\cos\chi + \ddot{h}_2\sin\chi \\
    &=\alpha_2(\boldsymbol{x},\chi) + B_2(u_1-u_2)
\end{split}
\label{eq:ddoty2}
\end{equation}
where $\alpha_2$ can be calculated from (\ref{eq:x_in_dot}) whereupon includes nonlinear terms and model uncertainties. $B_1$ and $B_2$ are control effectiveness on $y_1$ and $y_2$ respectively:

\begin{equation}
B_1 = - \bar{\kappa}R_{33,f}/m_v
\label{eq:hat_B1}
\end{equation}
\begin{equation}
\begin{split}
B_2 &= 
s_lh_{3,f}(G_p\sin\chi-G_q\cos\chi) \\
&=-\frac{h_{3,f}\bar{\kappa}b\sin\beta}{I_x\cos\zeta}\sin(\zeta-|\chi|)
\end{split}
\label{eq:hat_B2}
\end{equation}
where $\zeta$ is a positive virtual angle defined as
\begin{equation}
    \zeta = \tan^{-1}\left({\frac{I_x}{I_y}}\cot\beta\right)
    \label{eq:zeta_define}
\end{equation}

Now, from (\ref{eq:ddoty1}) and (\ref{eq:ddoty2}), the estimated control effectiveness matrix in (\ref{eq:control_INDI}) can be described as
\begin{equation}
    \boldsymbol{\mathcal{\hat{B}}}(\boldsymbol{x}_f)=
    \left[
    \begin{array}{cc}
        {B}_1 & {B}_1 \\
        {B}_2 & -{B}_2 
    \end{array}
    \right]
\end{equation} 
The estimation error of $\boldsymbol{\hat{\mathcal{B}}}$ mainly stems from the error of $m_v$, $I_x$, $I_y$, and $\bar{\kappa}$. Note that the filtered variables $R_{33,f}$ and $h_{3,f}$ are used in (\ref{eq:hat_B1}) and (\ref{eq:hat_B2}) because $\hat{\boldsymbol{\mathcal{B}}}$ is a function of $\boldsymbol{x}_f$ as per (\ref{eq:control_INDI}).

$Remark~\mathit{1}$: 
As indicated by (\ref{eq:ddoty2}) and (\ref{eq:hat_B2}), the system has the largest control effectiveness on $y_2$ when $|\sin(\zeta-|\chi|)|=1$. On the contrary, the control effectiveness becomes zero when $\sin(\zeta-|\chi|)=0$. Small control effectiveness leads to large control input command and subsequently deteriorates the control performance with the presence of actuator position and rate limit. Therefore, we enforce the effectiveness on $y_2$ to be greater than the minimum of $G_p$ and $G_q$, which yields the following constraint on $|\chi|$:
\begin{equation}
    r_B(|\chi|) \triangleq \frac{|{B_2}(|\chi|)|}{\min\{|G_p|, |G_q|\}} \geq 1 
    \label{eq:B2_chi_constraint}
\end{equation}
In addition, the following constraints are made to prevent ${B}_1 = {B}_2=0$, which is rather easy to fulfill:
\begin{itemize}
    \item $R_{33,f}\neq 0$: the thrust direction does not remain in the horizontal plane of $\mathcal{F_I}$.
    \item $h_{3,f} \neq 0$: $\boldsymbol{n}_d$ is not perpendicular to the current thrust direction ($-\boldsymbol{z}_B$).
\end{itemize}

\subsubsection{Second derivative of the output}
$\ddot{\boldsymbol{y}}_f=[\ddot{y}_{1,f},~\ddot{y}_{2,f}]^T$ in (\ref{eq:control_INDI}) can be obtained by directly taking the second-order derivative of filtered outputs. This, however, is prone to be detrimentally affected by measurement noise. Therefore, we can approximate $\ddot{y}_{1,f}$ by:
\begin{equation}
    \ddot{y}_{1,f} = \ddot{Z}_f = \dot{V}_{z,f} \simeq a_{z,f}{R}_{33,f} + g
\end{equation}
where $a_{z,f}$ is the projection of the filtered accelerometer measurement on $\boldsymbol{z}_B$.

$\ddot{y}_{2,f}$ can be obtained by numerically differentiating filtered $\dot{y}_2$. The latter can be derived from (\ref{eq:relaxed_att_kine_expand}) and (\ref{eq:output2}): 
\begin{equation}
 \dot{y}_2 = \cos\chi(-h_{3}q+h_{2}r+\lambda_1) + \sin\chi(h_3p-h_1r+\lambda_2)  
\end{equation}

\subsubsection{Pseudo-input definition}
The last step is to define the pseudo-input $\boldsymbol{\nu}$ as per (\ref{eq:nu_ndi}). As presented in (\ref{eq:ddoty1}) and (\ref{eq:ddoty2}), the control input $\boldsymbol{u}$ appears after taking the second derivative of both $y_1$ and $y_2$. Thus the system relative degrees are $\rho_1=\rho_2=2$. According to (\ref{eq:external_states_transformation1}) and (\ref{eq:external_states_transformation2}), there are four external states:
\begin{equation}
\begin{split}
&\left[\xi_{1},~\xi_{2},~\xi_{3},~\xi_{4}\right]^T
=\left[ y_1,~\dot{y}_1,~y_2,~\dot{y}_2\right]^T=\\
&\left[
\begin{array}{c}
     Z \\
     V_z \\
     h_1\cos{\chi} +  h_2\sin{\chi} \\
     (-h_3q+h_2r + \lambda_1)\cos{\chi} + (h_3p-h_1r+\lambda_2)\sin{\chi}
\end{array}\right]
\end{split}
\label{eq:external1}
\end{equation}
For this problem, the output reference is defined as
\begin{equation}
    \boldsymbol{y}_\mathrm{ref} = [Z_\mathrm{ref},~h_{1,\mathrm{ref}}\cos\chi+h_{2,\mathrm{ref}}\sin\chi]^T = [Z_\mathrm{ref},~0]^T
    \label{eq:output_control_DRF}
\end{equation}
Then, by substituting (\ref{eq:external1}) and (\ref{eq:output_control_DRF}) into (\ref{eq:nu_ndi}), we obtain the pseudo-input
\begin{equation}
\boldsymbol{\nu} = \left[
\begin{array}{c}
     -k_{z,p}(\xi_1-Z_\mathrm{ref})-k_{z,d}(\xi_2-\dot{Z}_\mathrm{ref})+\ddot{Z}_\mathrm{ref})  \\
    -k_{a,p}\xi_3 -k_{a,d}\xi_4 
\end{array}
\right]
\end{equation}
with positive gains $k_{z,p},~k_{z,d},~k_{a,p},~k_{a,d}$ to be tuned. 

Eventually, the control effectiveness $\boldsymbol{\mathcal{\hat{B}}}$, $\ddot{\boldsymbol{y}}$ and $\boldsymbol{\nu}$ is substituted into (\ref{eq:control_INDI}) to obtain $\boldsymbol{u}_\mathrm{indi}$. The rotor speed command of the remaining rotors can be subsequently calculated using (\ref{eq:u_define1}) or (\ref{eq:u_define2}).

\section{Stability Analysis of Internal Dynamics}
\label{sec:stability_analysis}
For the attitude/altitude inner-loop, internal states that need to be analyzed regarding their stability properties. As (\ref{eq:output2}) shows, the selection of $\chi$ is of great importance for influencing the internal dynamics, which will be elaborated in this section.

\subsection{Relaxed Trimming Equilibrium}
\begin{figure}
    \centering
    \includegraphics[scale = 0.5]{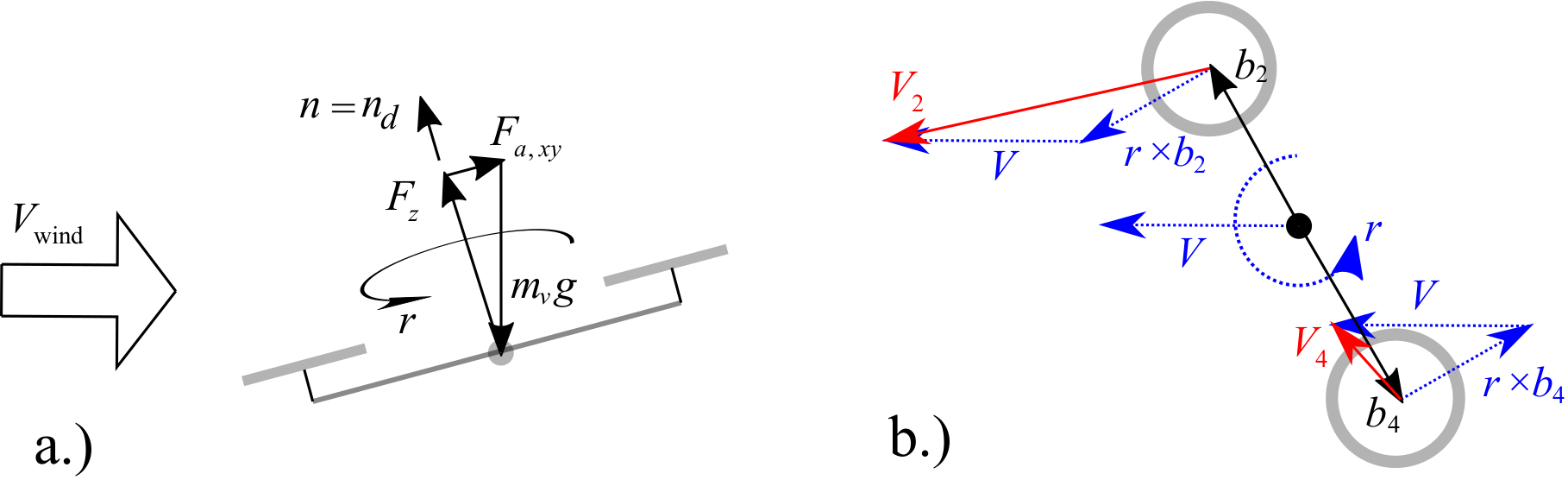}
    \caption{a.) Illustration of the force equilibrium at the relaxed trimming equilibrium. b.) The local velocities of the remaining rotors are different due to the high angular and translational speed of the drone (rotor 2 and 4 remain).}
    \label{fig:trimming_conditions_drawing}
\end{figure}

The relaxed trimming equilibrium is an extension of the relaxed hovering equilibrium~\cite{Mueller2015} to the high-speed flight regime where the aerodynamic drag becomes apparent. Note that the term trimming indicates the condition at a constant forward flight velocity. As Fig.~\ref{fig:trimming_conditions_drawing}a shows, the quadrotor spins about the axis $\boldsymbol{n}$ which represents the average thrust in a single revolution. In the relaxed trimming equilibrium, this averaged thrust is balanced with the average drag force (denoted by $\boldsymbol{F}_{a,xy}$) and the gravity.
If we assume the constancy of $\boldsymbol{M}_a$ and $\boldsymbol{F}_{a,z}$ in (\ref{eq:x_in_dot}), we have
\begin{equation}
    \boldsymbol{x}_\mathrm{in} = \bar{\boldsymbol{x}}_\mathrm{in} = [\bar{h}_1,~\bar{h}_2,~\bar{p},~\bar{q},~\bar{r},~\bar{Z},~\bar{V}_z]^T
    \label{eq:relaxed_trimming_condition}
\end{equation}
Specifically, if $\boldsymbol{n}^B = [0,~0,~-1]^T$, we have
\begin{equation}
\begin{array}{cc}
    \bar{h}_1 = \bar{h}_2 = \bar{p} = \bar{q} = 0
    \label{eq:relaxed_trimming_condition2}
\end{array}
\end{equation}
In practice, $\boldsymbol{M}_a$ and $\boldsymbol{F}_a$ are non-stationary. Thus variables $h_1$, $h_2$, $p$, $q$ and $r$ normally oscillate about the equilibrium. Nevertheless, as was analysed in \cite{Sun2018}, the average thrust direction remain unchanged as long as the reduced attitude $h_1$ and $h_2$ are bounded. The variation of yaw rate $r$ is also relatively small compared to $\bar{r}$. We therefore assume the constancy of $r$ in the following analysis.

Due to the spinning motion around the yaw axis during forward flight, the local airspeed and angle of attack of each rotor can be different (see Fig.~\ref{fig:trimming_conditions_drawing}b). The difference of local airspeed leads to the variation of thrust coefficient $\kappa$ of each remaining rotor \cite{Sun2019Aero}. The rotor speeds, therefore, vary with the heading angle and the variation grows with the flight speed.

\subsection{Internal Dynamics}
The internal dynamics are analyzed around the relaxed-trimming equilibrium. The following assumptions are further made to derive the internal states.
 
$Assumption~\mathit{1}$: $\boldsymbol{F}_a$ and $\boldsymbol{M}_a$ are independent from the control input $\boldsymbol{u}$.

$Assumption~\mathit{2}:$ The attitude reference $\boldsymbol{n}_d^I$ is slowly time-varying, thus $\dot{\boldsymbol{n}}_d^I=[\lambda_1,~\lambda_2,~\lambda_3]^T\simeq0$.

$Assumption~\mathit{3}: $ The average rotor speed $\bar{\omega}$ is considered as a constant that is independent from the control input $\boldsymbol{u}$. 

Note that $Assumption~\mathit{3}$ holds because of the near-constancy of the resultant thrust during the trimming condition, which has been verified from the flight data. But this assumption is invalid during aggressive thrust variation, such as vertical maneuvers. Nevertheless, the flight experiments given in Sec.~\ref{sec:experiment} will demonstrate that the internal dynamics are still stable during vertical maneuvers. 

Since $\boldsymbol{x}_\mathrm{in}$ has seven states in total and there are four external states as per (\ref{eq:external1}), we need to determine three internal states. The selection of internal states is not unique as long as the condition  (\ref{eq:phi_condition}) is satisfied, namely the first-order derivatives of $\boldsymbol{\eta}$ do not include $\boldsymbol{u}$. Based on the above assumptions, we hereby make the following choices for the internal states:
\begin{equation}
    \left[\begin{array}{c}
         \eta_1  \\
         \eta_2 \\
         \eta_3
    \end{array}\right] =
    \left[\begin{array}{c}
         -h_1\sin{\chi} +  h_2\cos{\chi}\\
         q\cos{\zeta}- s_lp\sin{\zeta}\\
         r+s_n\mu V_z
    \end{array}\right]
\label{eq:internal}
\end{equation}
where 
\begin{equation}
    \mu = {m_v\sigma}/{h_3}
\end{equation}
and $\zeta$ is defined as per (\ref{eq:zeta_define}). Note that the first internal state $\eta_1$ is the projection of $\boldsymbol{n}_d$ on $\boldsymbol{y}_S$ axis as illustrated in Fig.~\ref{fig:def_y2_chi}. Substituting (\ref{eq:relaxed_trimming_condition2}) into (\ref{eq:internal}) gives the internal state at relaxed hovering equilibrium $\bar{\boldsymbol{\eta}} = \boldsymbol{\phi}(\bar{\boldsymbol{x}}_\mathrm{in}) = [0,~0,~\bar{r}+s_n\mu\bar{V}_z]^T$.

$Proposition~\mathit{1}:$
The internal dynamics of the inner-loop system are locally asymptotically stable at the relaxed trimming equilibrium $\bar{\boldsymbol{x}}_\mathrm{in}$ if and only if $\chi = s_l|\chi|$ is selected such that every eigenvalue of the following $\boldsymbol{A}_1$ matrix has strictly negative real part:

\begin{equation}
    \boldsymbol{A}_1 = \frac{s_l}{\sin{(|\chi|-\zeta)}}\left[
    \begin{array}{cc}
      -\bar{r}\cos{(|\chi|-\zeta)}  &  1 \\
        -\bar{r}\Lambda& \Delta
    \end{array}
    \right]
    \label{eq:internal_dynamicsA}
\end{equation}
where 
\begin{equation}\Lambda = (A_x\bar{r}-2a_x\bar{\omega} s_n)\sin^2{\zeta}+(A_y\bar{r}+2a_y\bar{\omega} s_n)\cos^2{\zeta}\end{equation} 
\begin{equation}
\Delta =  -(A_x\bar{r} - 2 a_x\bar{\omega}s_n)\sin{\zeta}\sin{|\chi|}+(A_y\bar{r}+2 a_y\bar{\omega}s_n )\cos{\zeta}\cos{\chi}
\end{equation}

\begin{proof}
The transformation $[\boldsymbol{\eta},~\boldsymbol{\zeta}]^T=\boldsymbol{T}(\boldsymbol{x})$ expressed as (\ref{eq:external1}) and (\ref{eq:internal}) is a diffeomorphism if $h_3\sin(\zeta-|\chi|)\neq0$. The inverse transformation $\boldsymbol{x} =  \boldsymbol{T}^{-1}([\boldsymbol{\xi},~\boldsymbol{\eta}])$ thus can be obtained as

\begin{equation}
    \left[\begin{array}{c}
         Z\\
         V_z\\
         h_1\\
         h_2\\
         p\\
         q\\
         r
    \end{array}\right] =
    \left[\begin{array}{c}
         \xi_1 \\
         \xi_2\\
         \xi_3\cos\chi - \eta_1\sin\chi  \\
         \xi_3\sin\chi + \eta_1\cos\chi  \\
         \frac{h_3s_ns_l(\xi_4\cos\zeta+\eta_2\cos\chi)+(m_v\sigma \xi_2-\eta_3h_3)s_l\eta_1\cos\zeta}{h_3^2s_n\sin(|\chi|-\zeta)}\\
         \frac{h_3s_n(\xi_4\sin\zeta+\eta_2\sin|\chi|)+(m_v\sigma \xi_2-\eta_3h_3)\eta_1\sin\zeta}{h_3^2s_n\sin(|\chi|-\zeta)}\\
        (\eta h_3 - m_v\sigma \xi_2)/(h_3s_n)
    \end{array}\right]
    \label{eq:inv_T}    
\end{equation}
Then the dynamic equation of the internal states is derived as
\begin{equation}
         \dot{\boldsymbol{\eta}}= \tilde{\boldsymbol{f}_{\eta}}(\boldsymbol{x};\chi)
         = \tilde{\boldsymbol{f}_{\eta}}(\boldsymbol{T}^{-1}(\boldsymbol{\eta},\boldsymbol{\xi});\chi)
         =\boldsymbol{f}_{\eta}(\boldsymbol{\eta},\boldsymbol{\xi};\chi)
\end{equation}

A sufficient condition of the local stability of internal dynamics can be established via the notion of the zero dynamics \cite{Wallner2003AttitudeDynamics}:
\begin{equation}
    \dot{\boldsymbol{\eta}}= \boldsymbol{f}_{\eta}(\boldsymbol{\eta},0;\chi)
    \label{eq:internal_dyn_zero}
\end{equation}
By substituting (\ref{eq:inv_T}) into (\ref{eq:internal_dyn_zero}), we have:

\begin{equation}
    \dot{\eta}_1 = \frac{\eta_2 s_n h_3^2  - \eta_1 \eta_3 \cos(|\chi| - \zeta) h_3}{h_3 s_n s_l\sin(|\chi| - \zeta)}
    \label{eq:eta_dot1}
\end{equation}

\begin{equation}
         \dot{\eta}_2 =  \frac{
         \begin{array}{c}
              \sin\zeta (- 2 a_x \bar{\omega}  + A_x \eta_3) (\eta_1 \eta_3 \sin\zeta - \eta_2 h_3 s_n \sin|\chi|)  \\
              -\cos\zeta (2 a_y \bar{\omega}  + A_y \eta_3) (\eta_1 \eta_3 \cos\zeta - \eta_2 h_3 s_n \cos\chi) 
         \end{array}}{s_lh_3 \sin(|\chi| - \zeta)}\\
             \label{eq:eta_dot2}
\end{equation}

\begin{equation}
         \dot{\eta}_3 =  
         \begin{array}{c}
              g m_v \sigma/h_3 - s_n\eta_3\gamma/s_n  \\
              -\frac{s_lA_z (\eta_1 \eta_3 \cos\zeta - \eta_2 h_3 s_n \cos\chi) (\eta_1 \eta_3 \sin\zeta - \eta_2 h_3 s_n \sin|\chi|)}{h_3^2  \sin(|\chi| - \zeta)^2} 
         \end{array}
             \label{eq:eta_dot3}
\end{equation}

According to the first Lyapunov criterion, the equilibrium of the nonlinear system is asymptotically stable if the linearized system is asymptotically stable~\cite{Khalil2002NonlinearSystems}. 
At the relaxed trimming equilibrium, the internal states are $\bar{\boldsymbol{\eta}}=[0,~0,~\bar{r}+s_n\mu\bar{V}_z]^T$, and the local linearized system is derived from (\ref{eq:eta_dot1})-(\ref{eq:eta_dot3}) as 
\begin{equation}
         \left[
    \begin{array}{c}
         \dot{\eta}_1  \\
         \dot{\eta}_2  \\
         \dot{\eta}_3
    \end{array}
    \right]= 
    \left[
    \begin{array}{cc}
        \boldsymbol{A}_1 & \mathbf{O}_{2\times1} \\
        0 & -\gamma/I_z
    \end{array}
    \right]
    \Bigg(\left[
    \begin{array}{c}
         {\eta}_1  \\
         {\eta}_2  \\
         {\eta}_3 
    \end{array}
    \right]-\bar{\boldsymbol{\eta}}\Bigg)
    \label{eq:internal_dyn}
\end{equation}
where $\boldsymbol{A}_1$ is expressed as per (\ref{eq:internal_dynamicsA}). Note that the yaw damping $\gamma$ is positive definite and the system matrix of (\ref{eq:internal_dyn}) is block diagonal. Therefore, if $\boldsymbol{A}_1$ is Hurwitz, namely every eigenvalue of  $\boldsymbol{A}_1$ has strictly negative real part, then the linear system (\ref{eq:internal_dyn}) is asymptotically stable. Subsequently the local asymptotic stability of the internal dynamics is satisfied.
\end{proof}

One may approximate $\bar{r}$ and $\bar{\omega}$ from (\ref{eq:x_in_dot}) with $\boldsymbol{M}_a$ and $\boldsymbol{F}_a$ neglected:
\begin{equation}
    \bar{r} = -s_nm_vg \sigma/ \gamma,~\bar{\omega} = \sqrt{\frac{m_vg}{2\bar{\kappa}}}
    \label{eq:r_bar_w_bar}
\end{equation}
$Remark~\mathit{2}$: From (\ref{eq:internal_dynamicsA}) and (\ref{eq:r_bar_w_bar}) we have
\begin{equation}
\mathrm{Re}\left(\boldsymbol{\lambda}_{\boldsymbol{A}_1}(s_ls_n,~|\chi|)\right) = -\mathrm{Re}\left(\boldsymbol{\lambda}_{\boldsymbol{A}_1}(-s_ls_n,~|\chi|)\right)
\label{eq:A1andslsn}
\end{equation}
where $\boldsymbol{\lambda}_{\boldsymbol{A}_1}$ denotes eigenvalues of $\boldsymbol{A}_1$. Note that for a specific quadrotor, the value of $s_ls_n$ is identical under both failure scenarios indicated by (\ref{eq:u_define1}) and (\ref{eq:u_define2}). Therefore, we can further conclude from (\ref{eq:A1andslsn}) that $\boldsymbol{\lambda}_{\boldsymbol{A}_1}$, i.e. the stability property of internal dynamics, remain invariant despite the failure type if $|\chi|$ is fixed.
\subsection{Case Study: Selection of $|\chi|$}
As previous analysis presents, parameter $|\chi|$ need to be selected such that : (1) The matrix $\boldsymbol{A}_1$ in (\ref{eq:internal_dynamicsA}) is Hurwitz for stable internal dynamics. (2) Condition (\ref{eq:B2_chi_constraint}) is satisfied for an acceptable control effectiveness on $y_2$.

In this section, we conduct a case study on a specific type of quadrotor, a modified Parrot Bebop2, in the simulation to demonstrate the effect of $|\chi|$ on the overall controller performance. The inertial and geometric property of this quadrotor is listed in Table \ref{tab:Bebop2 parameters}. Without loss of generality, we assume rotor 2 and 4 are removed ($s_n = -1$, $s_l = 1$, $\chi=|\chi|$). Thus from (\ref{eq:r_bar_w_bar}), we have $
    \bar{r} = 26.4~\mathrm{rad/s}, ~~ \bar{\omega} = 1015~ \mathrm{rad/s}
$.

\begin{figure}
    \centering
    \includegraphics[scale=0.7]{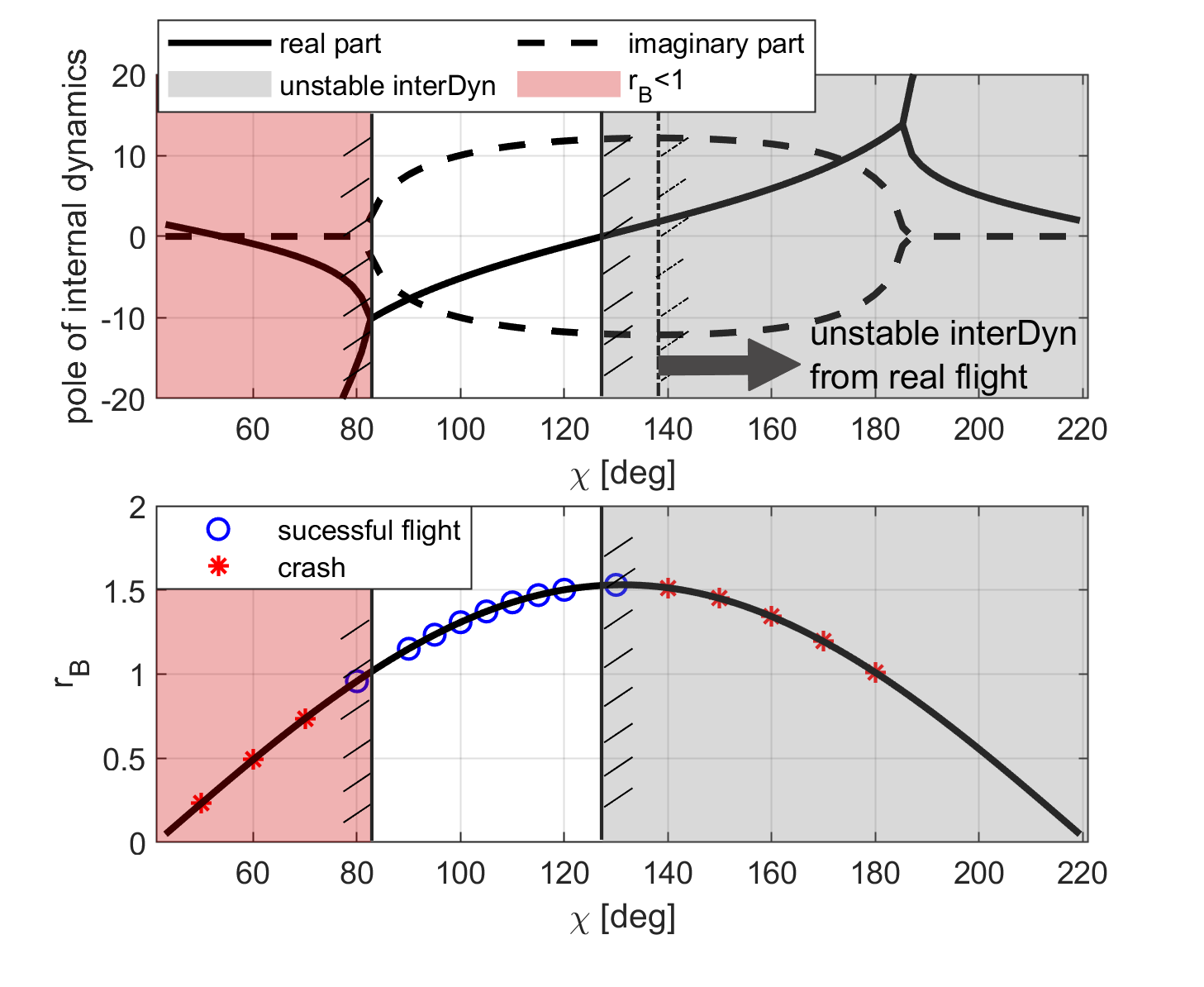}
    \caption{Poles of matrix $\boldsymbol{A}_1$ and $r_B$ varying with $|\chi|\in(\zeta,~\zeta+\pi)$. The unshadded area is the permitted region of $|\chi|$. Tests in the simulation are marked in the bottom plot with different symbols indicating if the flight succeeds. Unstable internal dynamic region obtained from real flight data in is shown in the top plot, which slightly moves rightward indicating a larger admissible region of $|\chi|$. Though $\boldsymbol{A}_1$ is stable in the area shaded red, crash still occurs because of $r_B<1$ that violates the constraint (\ref{eq:B2_chi_constraint}).}
    \label{fig:pole_rB_vs_chi}
\end{figure}

The top plot in Fig.~\ref{fig:pole_rB_vs_chi} shows the poles of $\boldsymbol{A}_1$ versus $|\chi|\in(\zeta,~\zeta+\pi)$. Note that $|\chi|=\zeta + k\pi$ ($k\in\mathbb{Z})$ causes singularity as (\ref{eq:internal_dynamicsA}) shows. The shadded gray area represents positive real part of poles that render unstable internal dynamics. The bottom plot in Fig.~\ref{fig:pole_rB_vs_chi} presents $r_B(|\chi|)$ as given in (\ref{eq:B2_chi_constraint}) with different $|\chi|$. The shaded red represents the violation of the constraint $r_B\geq1$. 

In the simulation, the quadrotor is commanded to transfer from $X = 0$ to $X=3~\mathrm{m}$ at $t=1~\mathrm{s}$. Various selections of $|\chi|$ are tested and given in the bottom plot of Fig.~\ref{fig:pole_rB_vs_chi}. The flights within the unshaded area succeed in conducting the transfer maneuver, whereas most of those in the shaded area failed. 

\begin{figure}
    \centering
    \includegraphics[scale=0.7]{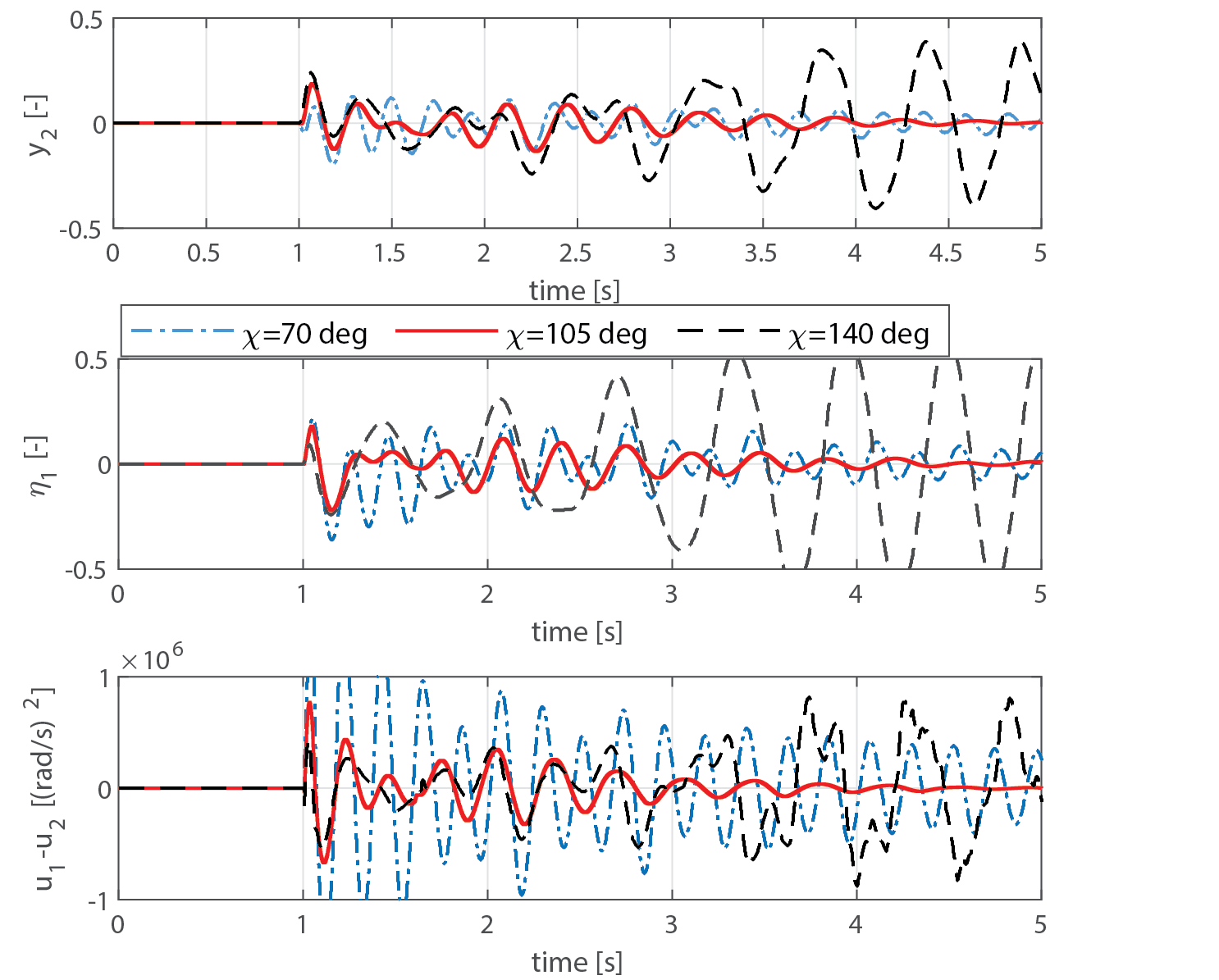}
    \caption{Time series of $y_2$, $\eta_1$, and $u_1-u_2$ of three tests with different $|\chi|$ in the simulation. A step-input of position command is given at $t=1$ s.}
    \label{fig:y2e1andU}
\end{figure}

Three tests in the simulation with respective $|\chi|$ equal to 70, 105, and 140 degrees are further demonstrated. Fig.~\ref{fig:y2e1andU} shows the time series of output $y_2$, internal state $\eta_1$, and $u_1-u_2$ of these three flights. When $|\chi| = 105~\mathrm{deg}$, the transition is successful where both $y_2$ and internal states $\eta_1$ converge to zero. As $|\chi|=70~\mathrm{deg}$, the violation of constraint $r_B\leq 1$ leads to a small control effectiveness $\hat{B}_2$. As a result, $u_1 - u_2$ significantly oscillates during the maneuver and the drone crashed due to limited actuator dynamics. On the other hand, when $|\chi|=140 ~\mathrm{deg}$, the internal dynamics are unstable and divergent oscillation of $\eta_1$ occurs that makes the drone crash.

\section{Generalization to Other Failure Conditions}
\label{sec:towradsAUnifiedApproach}
In this section, the inner-loop control scheme introduced in Sec.~\ref{sec:controller_design} is generalized to a quadrotor with complete loss of a single rotor, or without rotor failure (nominal condition).

\subsection{Single Rotor Failure}
\subsubsection{Internal Dynamics Analysis}
For a quadrotor with a single rotor failure, there are three permitted inputs. Therefore, three outputs can be defined. Similar to the condition with failure of two rotors, one output is defined as the altitude:
\begin{equation}
     y_1 = Z  \\
     \label{eq:output1SRF}
\end{equation}
The other two outputs are related to the reduced attitude control:
\begin{equation}
     y_2 = h_1 - n_x^B
\label{eq:output2SRF}
\end{equation}
\begin{equation}
    y_3 = h_2 - n_y^B
\label{eq:output3SRF}        
\end{equation}
To align the body fixed unit vector $\boldsymbol{n}^B = [n^B_x,~n^B_y,~n^B_z]^T$ with the reference $\boldsymbol{n}^B_d=[h_1,~h_2,~h_3]^T$, the preceding $y_2$ and $y_3$ should be stabilized to zero.

The relative degrees of the inner-loop system are $\rho_1 = \rho_2 = \rho_3 = 2$ from (\ref{eq:x_in_dot}). Therefore, the external states are

\begin{equation}
    \left[\begin{array}{c}
         \xi_1\\
         \xi_2\\
         \xi_3\\
         \xi_4\\
         \xi_5\\
         \xi_6
    \end{array}\right]=
    \left[\begin{array}{c}
         y_1\\
         \dot{y}_1\\
         y_2\\
         \dot{y}_2\\
         y_3\\
         \dot{y}_3
    \end{array}\right]=
    \left[\begin{array}{c}
        Z  \\
        V_z\\
        h_1-n_x^B\\
        -h_3q+h_2r\\
         h_2-n_y^B\\
         h_3p-h_1r
    \end{array}\right]   
\end{equation}
Note that $\boldsymbol{x}_\mathrm{in}\in\mathbb{R}^7$ and there are 6 external states, the only internal state can be selected as
\begin{equation}
\eta_1 = r + \mu_1 V_z + \mu_2 p + \mu_3 q
\end{equation}
where $\mu_1$, $\mu_2$, $\mu_3$ are calculated such that $\dot{\eta}_1$ does not include the control input. These coefficients are constant and related to the handedness of the remaining rotors. 

After some tedious algebra, the zero dynamics can be calculated as \begin{equation}
    \dot{\eta}_1 = -\frac{\gamma}{\Theta}\eta_1 + \frac{\Pi}{h_3\Theta^2}\eta_1^2 + g\mu_1
    \label{eq:zerodynSRF}
\end{equation}
where 
\begin{equation}
    \Theta = n_x^B\mu_2/h_3 + n_y^B\mu_3/h_3 + 1
\end{equation}
\begin{equation}
    \Pi = A_zn_x^Bn_y^B/h_3 + A_xn_y^B\mu_2 + A_yn_x^B\mu_3
\end{equation}

Particularly, when $n_x^B = n_y^B = 0$, namely the drone spins about it thrust direction, we have
\begin{equation}
\dot{\eta}_1 = -\gamma(\eta_1-\bar{\eta}_1)
\end{equation}
Since the yaw damping $\gamma > 0$, the internal state $\eta_1$ at the equilibrium $\bar{\eta}_1$ is stable.   
\subsubsection{Control Law}
The above analysis demonstrates that the stability of internal dynamics with the selection of outputs in (\ref{eq:output1SRF})-(\ref{eq:output3SRF}). After selecting the control outputs, the same control scheme presented in Fig.~\ref{fig:control_scheme_DRF} can be applied for the single rotor failure condition. Without loss of generality, we assume that rotor 4 is removed, then we have
\begin{equation}
    \boldsymbol{u} = [\omega_1^2,~\omega_2^2,~\omega_3^2]^T
\end{equation}
With the same procedure introduced in Sec.~\ref{sec:controller_design}, the control law for a quadrotor subjected to a single rotor failure can be obtained using (\ref{eq:control_INDI}), where

\begin{equation}
\boldsymbol{\nu} = \left[
\begin{array}{c}
     -k_{z,p}(\xi_1-Z_\mathrm{ref})-k_{z,d}(\xi_2-\dot{Z}_\mathrm{ref})+\ddot{Z}_\mathrm{ref})  \\
     -k_{a,p}\xi_3-k_{a,d}\xi_4\\
     -k_{a,p}\xi_5-k_{a,d}\xi_6\\
\end{array}
\right]
\end{equation}
\begin{equation}
{\boldsymbol{y}}_f^{(\boldsymbol{\rho})} = \ddot{\boldsymbol{y}}_f = [\dot{a}_{z,f}/R_{33,f},~\ddot{h}_{1,f},~\ddot{h}_{2,f}]^T
\end{equation}
The control effectiveness matrix $\hat{\boldsymbol{\mathcal{B}}}$ can be estimated using (\ref{eq:F}) and (\ref{eq:M}):
\begin{equation}
    \boldsymbol{\mathcal{\hat{B}}}(\boldsymbol{x}_f) = \left[\begin{array}{c c c}
- \bar{\kappa}R_{33,f}/m_v & - \bar{\kappa}R_{33,f}/m_v & - \bar{\kappa}R_{33,f}/m_v\\
-\bar{\kappa}b\sin\beta & \bar{\kappa}b\sin\beta & \bar{\kappa}b\sin\beta\\
\bar{\kappa}b\cos\beta & \bar{\kappa}b\cos\beta & -\bar{\kappa}b\cos\beta 
\end{array}
    \right]
\end{equation}
 
\subsection{Without Rotor Failure}
For a multi-rotor drone with more than three actuators, such as a nominal quadrotor or a hexacopter, there are four or more permitted control inputs. We can then introduce the fourth output related to the yaw control:
\begin{equation}
    y_4 = r
    \label{eq:NF_output}
\end{equation}
An independent yaw controller can be appended to provide the reference yaw rate $r_\mathrm{ref}$, such as a PD controller:
\begin{equation}
    r_\mathrm{ref} = -k_{p,\psi}e_{\psi} - k_{d,\psi}\dot{e}_{\psi}
\end{equation}
where $e_{\psi}$ is the yaw angle tracking error; $k_{p,\psi}$ and $k_{d,\psi}$ are positive gains.
 
Note that the rotor angular acceleration $\dot{\omega}_i$ may deteriorate the yaw control performance while implementing this approach. Interested readers are referred to \cite{Smeur2016adaptive} to tackle this problem for a nominal quadrotor.

\section{Experimental Validation} 
\label{sec:experiment}

\begin{table}[t]
  \caption{Parameters of the tested quadrotor.}
  \label{tab:Bebop2 parameters}
  \begin{center}
  \renewcommand{\arraystretch}{1.2}
  \resizebox{.45\textwidth}{!}{
  \begin{tabular}{ccccccc}
  \hline\hline
   par. & value & unit & & par. & value & unit \\ 
   \cline{1-3}\cline{5-7}
   $I_{x}$ &1.45$e^{-3}$ & kg$\cdot$ m$^2$ & & $m_v$ & 0.410 & kg\\
   $I_{y}$ &1.26$e^{-3}$ & kg$\cdot$ m$^2$ & & $b$ & 0.145 & m \\
   $I_{z}$ &2.52$e^{-3}$ & kg$\cdot$ m$^2$ & & $\beta$ & 52.6 & deg\\
   $I_{p}$ &8.00$e^{-6}$ & kg$\cdot$ m$^2$ & & $\gamma$ &1.50$e^{-3}$ & N$\cdot$m$\cdot$s \\
   $\bar{\kappa}$ &1.90$e^{-6}$ & kg$\cdot$ m$^2$ & & $\sigma$ & 0.01 & m\\
  \hline
  \end{tabular}}
  \label{tab:Bebop2 parameters}
  \end{center}
  \end{table}
  
The proposed control method has been validated on a modified Parrot Bebop2 drone with a lighter battery and camera module removed. The geometric and moment of inertia properties are given in Table~\ref{tab:Bebop2 parameters}. During the flight test, a motion capturing system (OptiTrack) with 12 cameras provided the position measurements of the 4 markers attached to the drone at 120~Hz. The inertial measurement unit (IMU) measured the angular rates (from gyroscope) and the specific force (from the accelerometer) at 512~Hz. A built-in brushless DC (BLDC) motor controller controled the rotor speeds of each propeller, and also measured the rotor rotational rates in RPM at 512~Hz. Subsequently, an Extended Kalman Filter (EKF) was implemented to estimate the position, velocity of the center of mass, and the attitude of the body frame. The proposed controller and the EKF were run onboard at 500~Hz with the original processor Parrot P7 dual-core CPU Cortex 9. The control gains are given in Table~\ref{tab:gains}. 

\begin{table}[]
    \caption{Control gains.}
    \centering
    \renewcommand{\arraystretch}{1.3}
    \resizebox{.45\textwidth}{!}{
    \begin{tabular}{lccccc}
    \hline\hline
         outer& $k_p~[s^{-2}]$ & $k_i~[s^{-3}]$ & $k_d~[s^{-1}]$ \\
         \cline{2-4}
         -loop&1.0& 0.1 & 1.0\\
         \hline
    inner& $k_{a,p}~[s^{-2}]$ & $k_{a,d}~[s^{-1}]$ & $k_{z,p}~[s^{-2}]$ &$k_{z,d}~[s^{-1}]$ \\
         \cline{2-5}
         -loop&50 &30 & 15 & 10\\
    \hline
    \end{tabular}}
    \label{tab:gains}
\end{table}
\subsection{Flight tests in windless conditions}

\begin{figure}
    \centering
    \hspace*{-2mm}%
    \includegraphics[scale = 0.75]{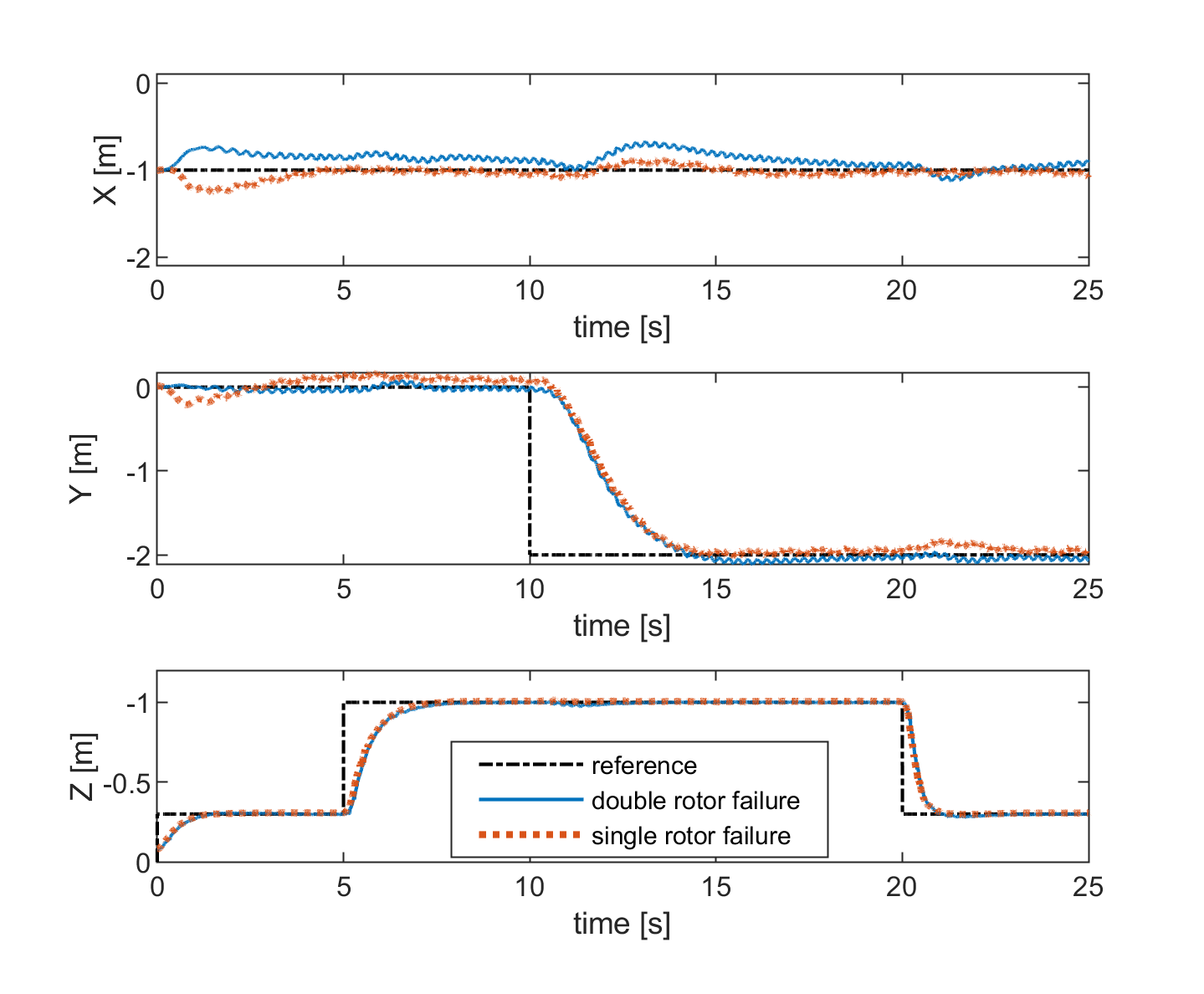}
    \caption{Position tracking task under windless condition. Blue solid lines indicate the 3-D position of the tested quadrotor with failure of two opposing rotors. Red dash lines indicate those under single rotor failure conditions. The reference positions are presented as black dot dash lines.}
    \label{fig:pos_track_srf_drf}
\end{figure}
\begin{figure}
    \centering
    \hspace*{0mm}%
    \includegraphics[scale = 0.75]{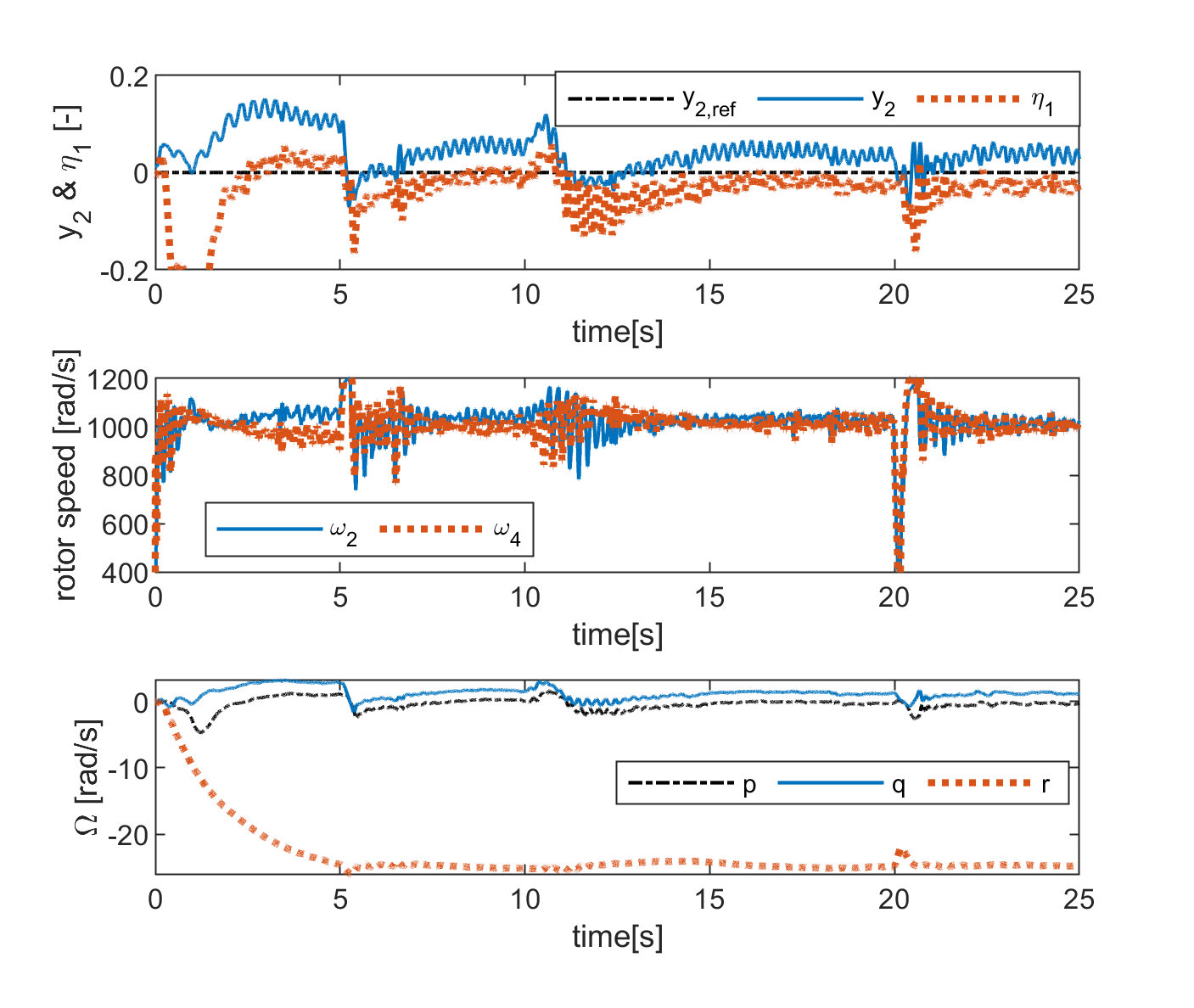}
    \caption{Variables during the position tracking flight test with failures of rotor 1 and rotor 3. From top to bottom are: the output $y_2$ and internal state $\eta_1$; the angular speed measurements of the rotor 2 and rotor 4; the angular rates measurements.}
    \label{fig:drf_other_states}
\end{figure}
\begin{figure}
    \centering
    \includegraphics[scale = 0.73]{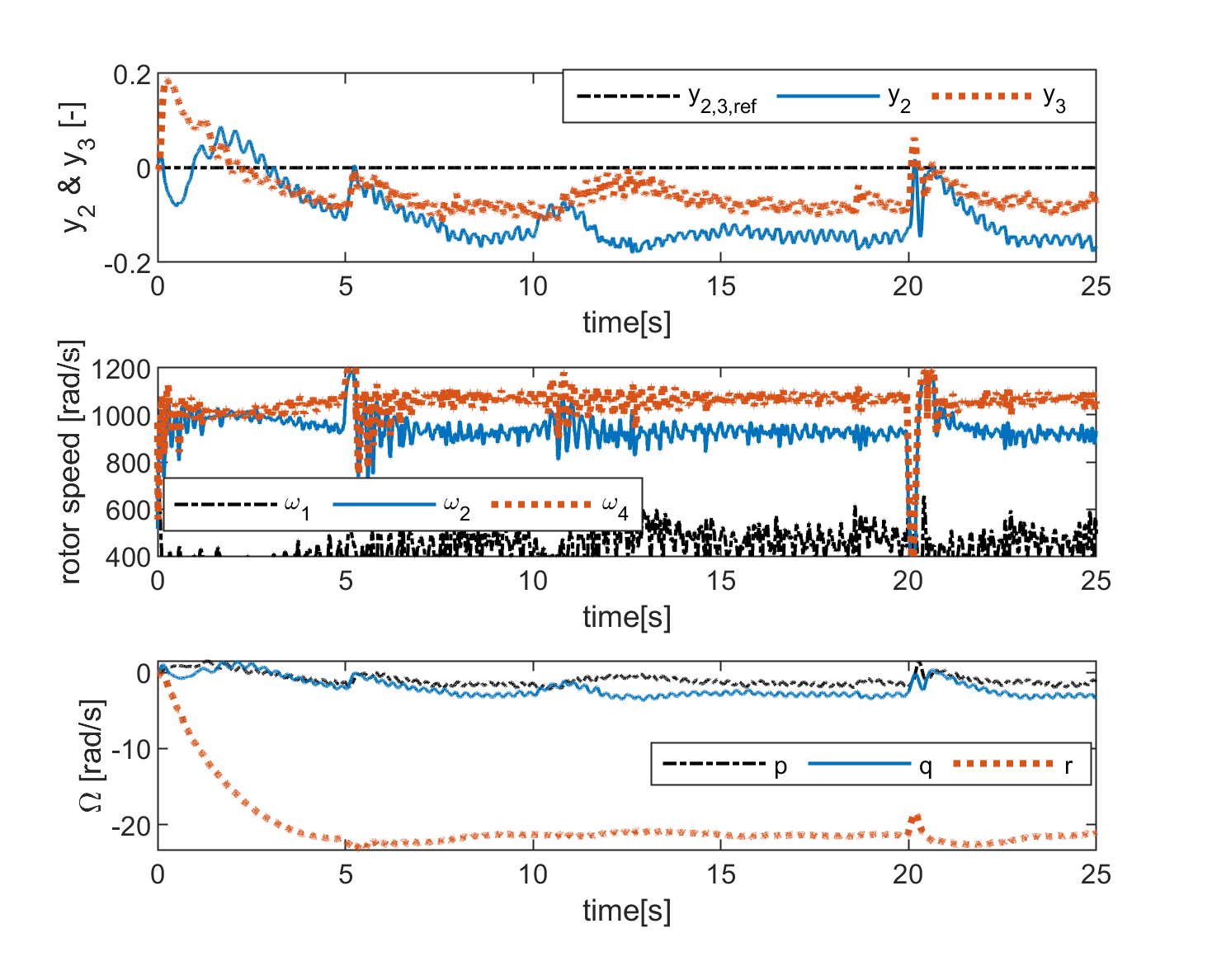}
    \caption{Variables during the position tracking flight test with failure of rotor 3. From top to bottom are: the output $y_2$ and $y_3$; the angular speed measurements of the rotor 1, 2, and 4; the angular rates measurements.}
    \label{fig:srf_other_states}
\end{figure}

The first set of flights tested the 3-D trajectory tracking of the quadrotor without wind disturbance. Fig.~\ref{fig:snapshot} presents snapshots of the tested quadrotor with failure of one and two rotors within 0.3~s; Fig.~\ref{fig:pos_track_srf_drf} shows the reference and the measured positions.

For those with two rotor failures, we removed the rotor 1 and rotor 3, without loss of generality. Fig.~\ref{fig:drf_other_states}a shows the second output $y_2$ and the first internal state $\eta_1$. As is introduced in Sec.~\ref{sec:controller_design}, they represent the reduced attitude and need to converge to zero. Despite the misalignment during the take-off maneuver at the first 3 seconds, a slight tracking error of $y_2$ is observed which is presumably due to the bias of the center of mass. In this flight, $|\chi|=90$~deg was selected for stable internal dynamics. Consequently, the internal state $\eta_1$ was confined around zero. Fig.~\ref{fig:drf_other_states}b shows the angular speeds of rotor 2 and 4 that remained almost constant at $\bar{\omega} = 1000$~rad/s during the horizontal maneuvers, which was in-line with the $Assumption~3$. A yaw rate at about $-25$~rad/s shown in Fig.~\ref{fig:drf_other_states}c indicates the fast spinning motion of the damaged quadrotor. 

With the same controller and the same set of gains, the condition with one rotor failure was also tested. The rotor 3 was removed in this test. As is shown in Fig.~\ref{fig:srf_other_states}a, the internal state $\eta_1$ is replaced by the third output $y_3$ because of the addition of one rotor compared to the condition with failure of two rotors. The reference $y_{2,\mathrm{ref}}=y_{3,\mathrm{ref}}=0$ was employed in this flight that required the rotor 1 (the one diagonal to the failed rotor) to generate no force. However, due to the lower saturation of rotor 1 presented in Fig.~\ref{fig:srf_other_states}b, a constant tracking error of $y_2$ and $y_3$ are observed. In spite of these attitude tracking errors, the drone under both failure cases were able to track the position commands.

\subsection{Effect of $\chi$ in the Condition with Failure of Two Opposing Rotors}
\begin{figure}
    \centering
    \includegraphics[scale=0.85]{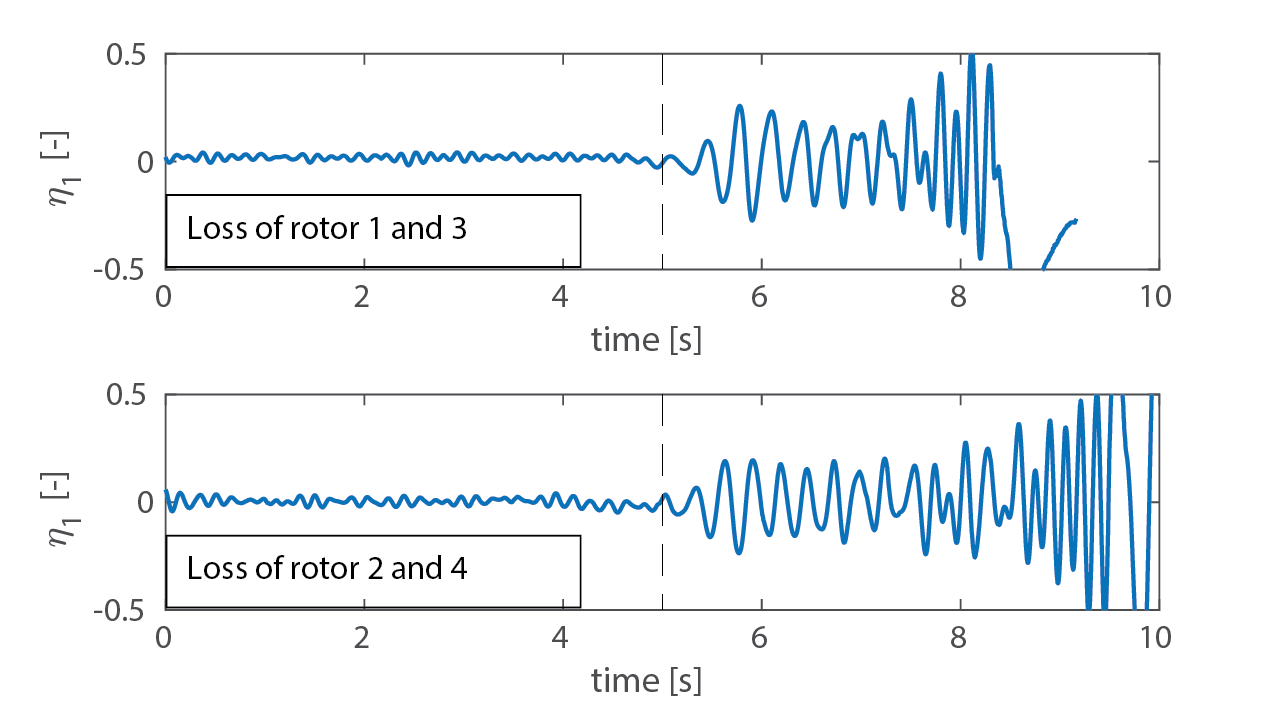}
    \caption{Internal state $\eta_1$ in with different selection of $|\chi|$. Before $t=5$~s, $|\chi| = 90$~deg is selected which leads to stable internal dynamics in both conditions. After $t=5$~s, $|\chi|$ is changed to $180$~deg and the internal dynamics become unstable.}
    \label{fig:drf_various_chi}
\end{figure}
The quadrotor in the condition with failure of two opposing rotors was tested in the hover region with different sets of $|\chi|$ to experimentally demonstrate its effect on the stability of the internal dynamics. Fig.~\ref{fig:drf_various_chi} shows the internal state $\eta_1$ during a hovering flight where
the parameter $|\chi|$ was initialized at 90~deg which would lead to stable internal dynamics. At $t=5$~s, $|\chi|$ was changed to $180$~deg during the flight and the internal state became unstable. This complies with the prediction from $Proposition$ $\it{1}$.

It is noteworthy that the stable region boundary moves slightly to the right compared to the theoretical prediction, as is shown in Fig.~\ref{fig:pole_rB_vs_chi}. As a consequence, the admissible region of $|\chi|$ becomes larger. The difference might come from the omission of the aerodynamic damping on pitch and roll rate while conducting internal dynamic analysis.

\section{Validations in a Wind Tunnel}
\label{sec:windtunnel}
To validate the robustness against unmodeled aerodynamic forces and moments in the high-speed flight regime, flight tests have been carried out in the Open Jet Facility (OJF), a large scale wind tunnel with an aperture of 2.85 meters (see Fig.~\ref{fig:drone_OJF}). The parameter $|\chi| = 105~\mathrm{deg}$ that lies roughly in the center of its admissible region is selected, as Fig.~\ref{fig:pole_rB_vs_chi} illustrates.

The linear quadratic regulator (LQR) is selected as the baseline for comparison in the wind tunnel. This method has been validated in practice in~\cite{Mueller2015}. The same set of gains from this paper were implemented for comparison. Specifically, the costs on control inputs were set to one, with units N$^{-2}$; the cost on the reduced attitude was set to 20 and the cost on angular rates was set to zero. Since a different drone was used, the time constant of the first-order actuator model is set as 30~ms that differs from \cite{Mueller2015}. Both INDI and LQR used the same outer-loop control gains as given in Table.~\ref{tab:gains}.

\subsection{Trajectory Tracking Task}
\begin{figure}
    \centering
    \hspace*{0mm}%
    \includegraphics[scale=0.85]{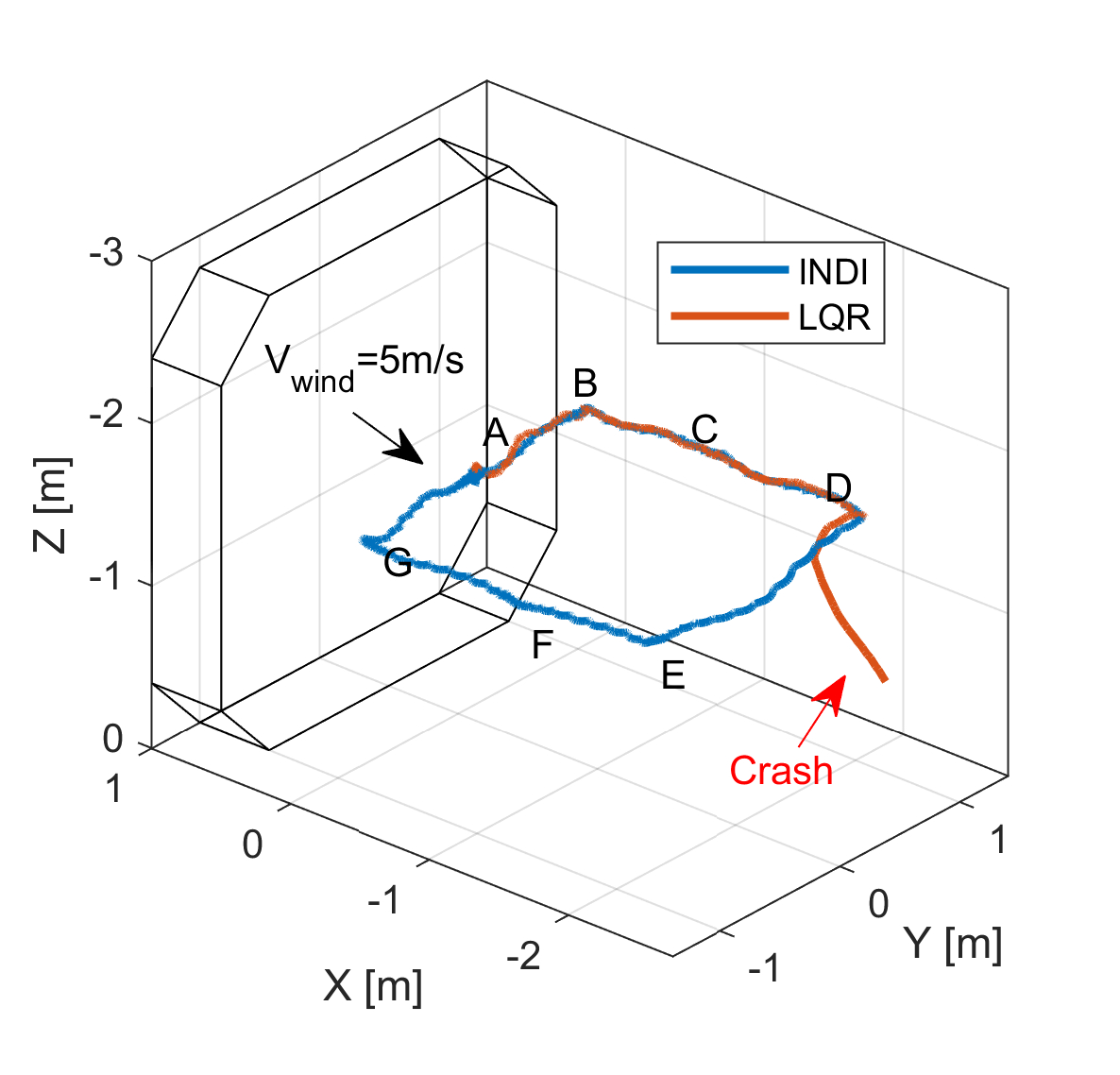}
    \caption{3D trajectories of the damaged quadrotor under $V_\mathrm{wind}=5$~m/s, where A to G represent the setpoints. INDI finished the trajectory tracking task while LQR failed during the transition from setpoint D to E.}
    \label{fig:3D_traj_windtunnel}
\end{figure}

\begin{figure}
    \centering
    \includegraphics[scale=0.75]{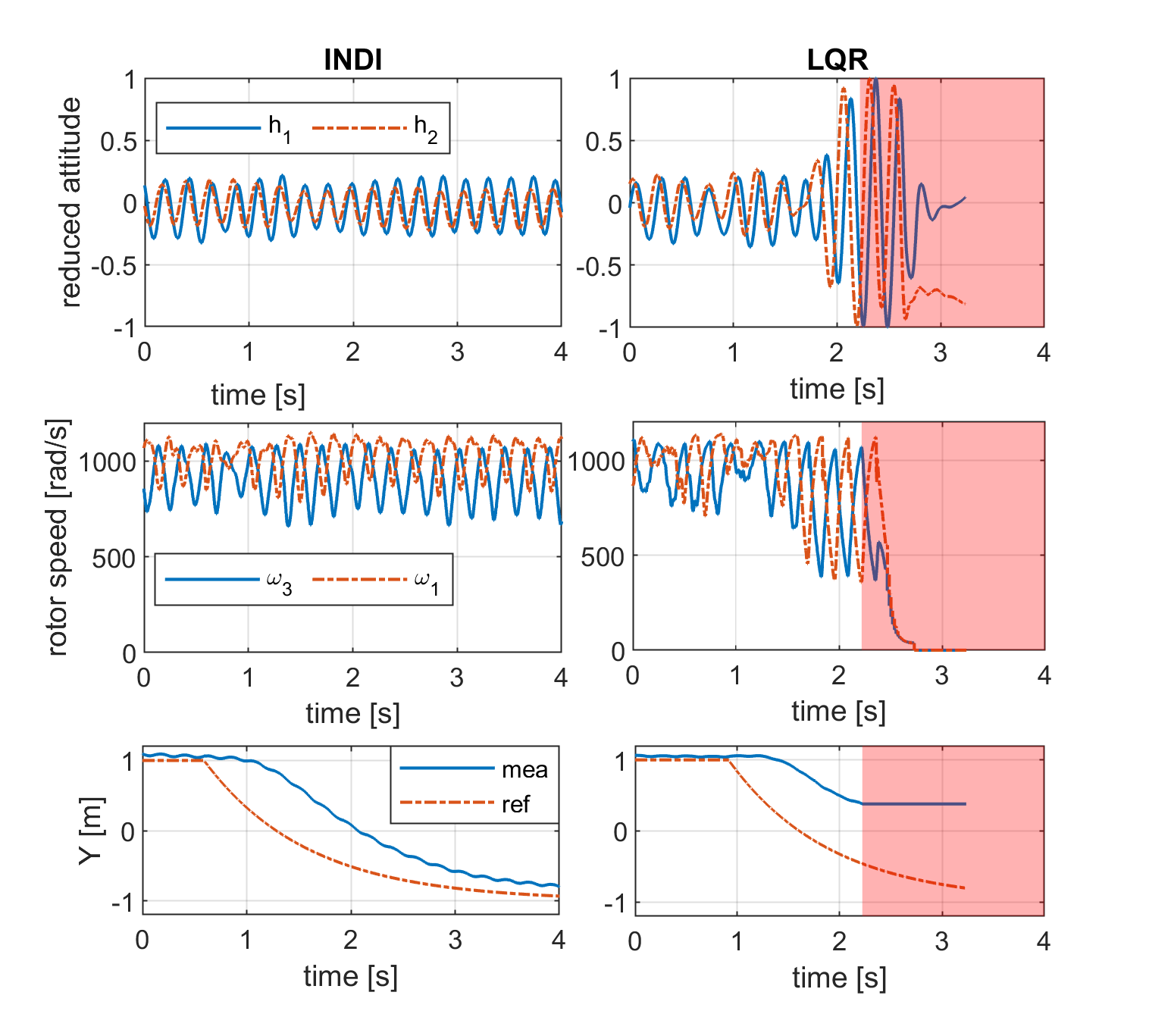}
    \caption{Reduced attitude, rotor speeds and lateral position $Y$ of the quadrotor during the transition from setpoint D to E. The reduced attitude of the drone controlled by LQR became unstable, before the loss of OptiTrack measurement shaded in red, and eventually crashed. In comparison, the drone controlled by the INDI approach succeed to finish this maneuver.}
    \label{fig:3D_traj_windtunnel_state}
\end{figure}

\begin{table}
  \caption{Maximum flight speed of the INDI and LQR controller with various control gains. $Q$ for LQR indicates the cost on the reduced attitude.}
  \label{tab:indi_vs_lqr}
  \begin{center}
  \renewcommand{\arraystretch}{1.2}
  \renewcommand{\thefootnote}{\fnsymbol{footnote}}
  \resizebox{.43\textwidth}{!}{
  \begin{tabular}{cccccc}
  \hline\hline
   \multicolumn{3}{c}{INDI} &   & \multicolumn{2}{c}{LQR} \\ 
   \cline{1-3} \cline{5-6}
   $k_{a,p}~[s^{-2}]$ & $k_{a,d}~[s^{-1}]$ & $V_\mathrm{max}$ [m/s] & & $Q~[-]$ & $V_\mathrm{max}$ [m/s] \\ 
      \cline{1-3} \cline{5-6}
   5 & 1 &7.8 & & 1 & 4.6 \\
   10 & 2 &8.3 & & 3 & 5.9\\
   $~50^{*}$ & $~30^{*}$ &$8.8$ & & 10 & 5.2 \\
   100 & 30 &8.2 & & $~20^{*}$ & 5.1 \\
   200 & 50 &7.8 & & 30 & 6.3 \\   
  \hline
  \multicolumn{6}{l}{*Gains for the trajectory tracking task.}
  \end{tabular}}
  \label{tab:Bebop2 parameters}
  \end{center}
  \end{table}

A trajectory tracking task was performed under a wind flow of 5~m/s. The wind flow was along the negative direction of the $\boldsymbol{x}_I$ axis. Fig.~\ref{fig:3D_traj_windtunnel} plots the trajectories using INDI and LQR respectively. The drone tracked setpoints A to G in sequence every 3 seconds. In addition, the step reference positions ($X_\mathrm{ref}$ and $Y_\mathrm{ref}$) were smoothed by a first-order filter with a time constant 1~s. As Fig.~\ref{fig:3D_traj_windtunnel} shows, both methods successfully tracked the setpoints before point D. However, the LQR approach failed during the transition from setpoint D to E, which was perpendicular to the wind flow. 

Fig.~\ref{fig:3D_traj_windtunnel_state} compares states between the two approaches in this period. From the plotted data, we can find that the reduced attitudes of both methods were bounded before conducting the maneuver, despite oscillations of reduced attitude and rotor speeds caused by the wind flow and yaw motion. During the maneuver from point D to E, the reduced attitude of the LQR approach diverged from zero (the linearization point). The instability of the inner-loop attitude control caused the failure of trajectory tracking and eventually lead to a crash. 

For LQR, the rotor speed commands (i.e. the control input) were naturally amplified (at around $t=2$~s in the mid-right plot of Fig.~\ref{fig:3D_traj_windtunnel_state}) to stabilize the reduced attitudes ($h_1$ and $h_2$) as they were diverging from the equilibrium (top-right plot). These increase of control input may destabilize the system in the presence of nonlinearities and model uncertainties caused by the wind flow. By contrast, the INDI approach used the second-order derivative of the reduced attitude (${\boldsymbol{y}}_{f}^{(\boldsymbol{\rho})}$ in (\ref{eq:control_INDI})) to compensate for the model uncertainties. The nonlinearities are also handled by the feedback linearization step of INDI. Thereby the reduced attitude can be stabilized without drastically increasing the control input.

\subsection{Maximum Flight Speed Test}
\begin{figure}[t!]
    \centering
    \includegraphics[scale=0.80]{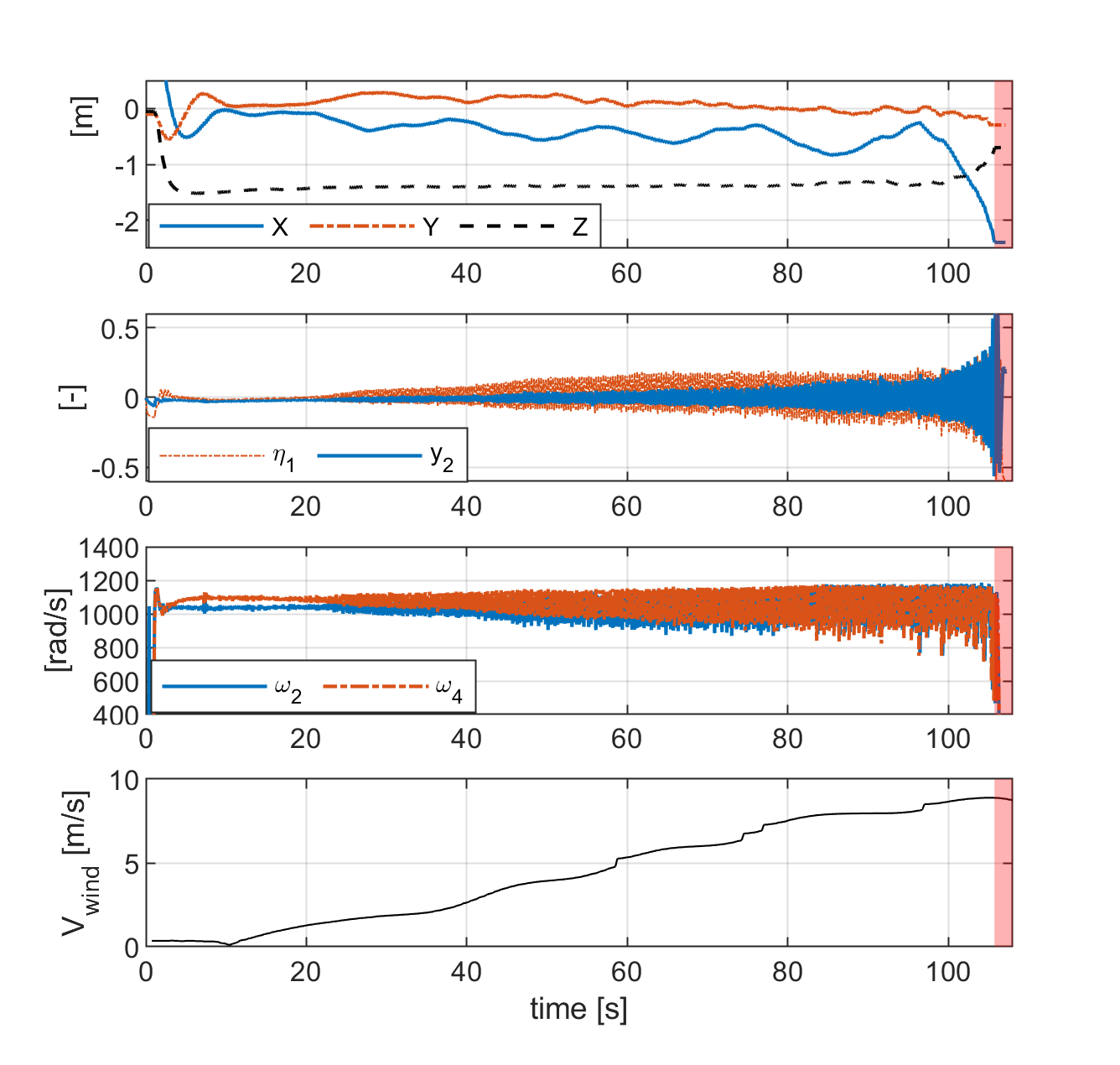}
    \caption{Variables during the wind-tunnel maximum flight speed test of INDI. From top to bottom are: positions of the drone; the internal state $\eta_1$ and the output $y_2$; the angular speed of the remaining rotor (rotor 2 \& 4); the time series of the wind speed which gradually increased until the loss-of-control happened. The red area represents loss of OptiTrack measurement.}
    \label{fig:variables_windtunnel}
\end{figure}
\begin{figure}[t!]
    \centering
    \includegraphics[scale=0.80]{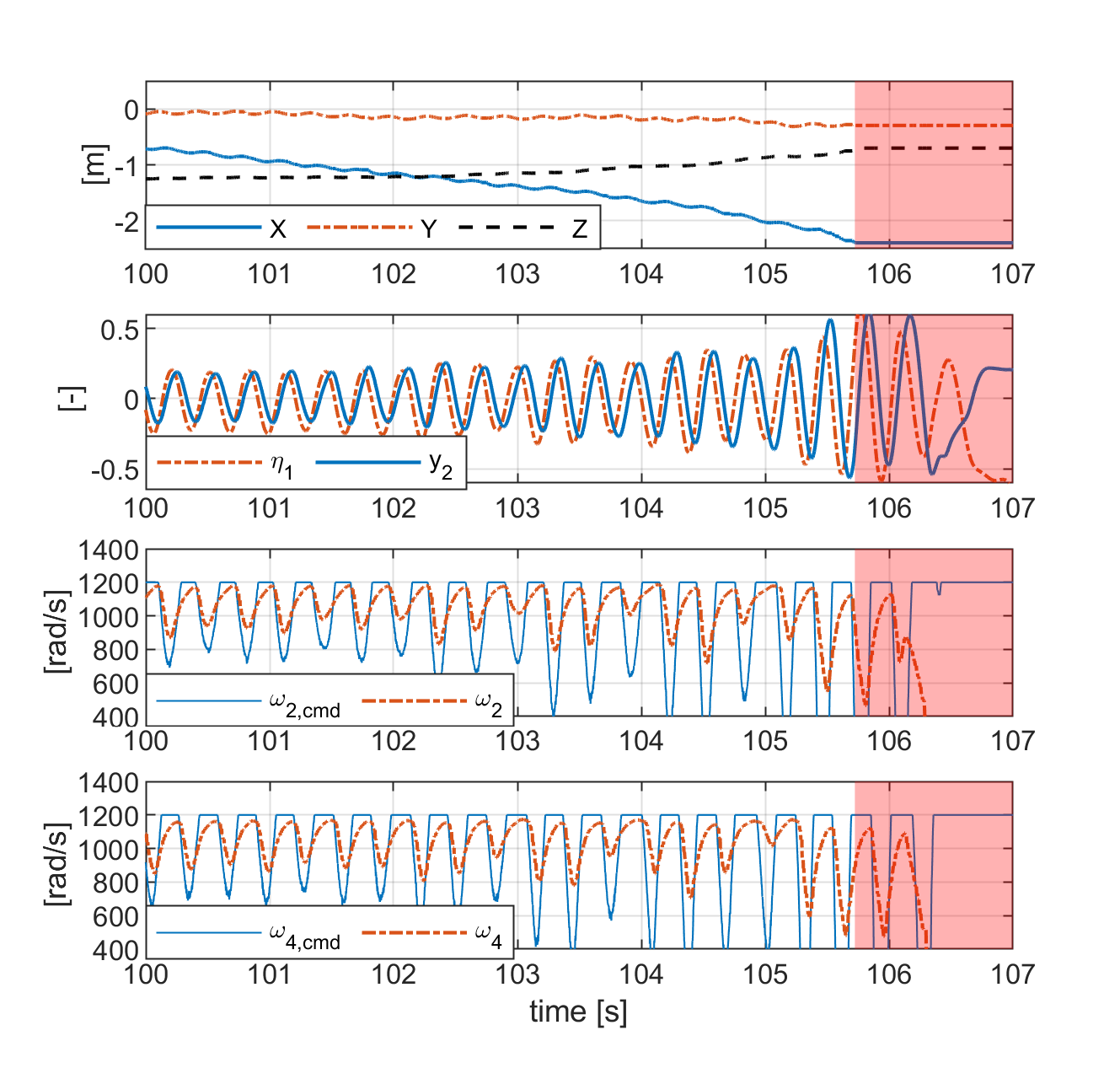}
    \caption{Close-up of the variables of Fig.~\ref{fig:variables_windtunnel} during last 7 seconds before the crash. The red area represents loss of OptiTrack measurement.} 
    \label{fig:variables_windtunnel_amplified}
\end{figure}
To explore the maximum flight speed of a quadrotor under the failure of two opposing rotors, maximum speed flight tests were conducted using both INDI and LQR. During the entire flight, the drone was hovering at a setpoint located in front of the wind tunnel (${\boldsymbol{P}_\mathrm{ref}} = [0,~0,~-1.5]^T$).
The wind speed was gradually increased from $V_\mathrm{wind}=0$ to a $V_\mathrm{max}$ until the drone crashed.
 
Table.~\ref{tab:indi_vs_lqr} compares the maximum flight speed achieved by the two approaches with different set of gains. Similarly to the trajectory tracking task in the preceding section, INDI outperforms the LQR approach in terms of maximum flight speed.
Be that as it may, the drone controlled by LQR was still stabilized at relatively high-speeds thanks to the inherent stabilizing property of feedback control. 

Fig.~\ref{fig:variables_windtunnel} shows the time series of position, $\eta_1$, $y_2$, rotor speeds and wind speed during a flight controlled by INDI, where the drone crashed at $V_\mathrm{wind}=8.8$~m/s when $t=107$~s. The variations of $\eta_1$, $y_2$ and rotor speeds significantly increased with the wind speed. Despite the gradual increase of the oscillation, the internal state $\eta_1$ was bounded near zero and subsequently ensured successful position tracking.

To investigate the cause of loss-of-control of INDI, Fig.~\ref{fig:variables_windtunnel_amplified} shows the close-up of Fig.~\ref{fig:variables_windtunnel} during the last 7 seconds before the crash. In addition to the rotor speed measurements, the rotor speed commands are also plotted. From the reduction of $X$ in the top plot, we can find that the quadrotor was blown away from the setpoint along the wind flow. Meanwhile, the increase of $Z$ indicates the continues reduction of the altitude in this process. These phenomena are believed caused by the saturation of motors under wind resistance, which can be clearly seen in the bottom two plots of Fig.~\ref{fig:variables_windtunnel_amplified}. In addition to the motor saturation, the control input lag due to motor dynamics can be observed, which may cause the gradual divergence of $\eta_1$ ad $y_2$ in the second plot of Fig.~\ref{fig:variables_windtunnel_amplified}. We hence infer that the motor capacity (bandwidth and power limit) is a major limiting factor of the maximum flight speed.

\subsection{High-Speed Flight with Imperfect State Estimations}
Since the ultimate goal of this work is improving drone safety during the high-speed flight in outdoor environments, preliminary validations of the proposed method using imperfect state estimations have been conducted in the wind-tunnel. The sampling rate of the motion capturing system was reduced from 120~Hz to 10~Hz to simulate GPS-like update rates. Only position measurements were transmitted to the onboard flight controller. A complementary filter~\cite{Mahony2008} was implemented by fusing the measurements from the IMU and the magnetometer, to provide attitude estimates. 

\begin{figure}
    \centering
    \includegraphics[scale=0.85]{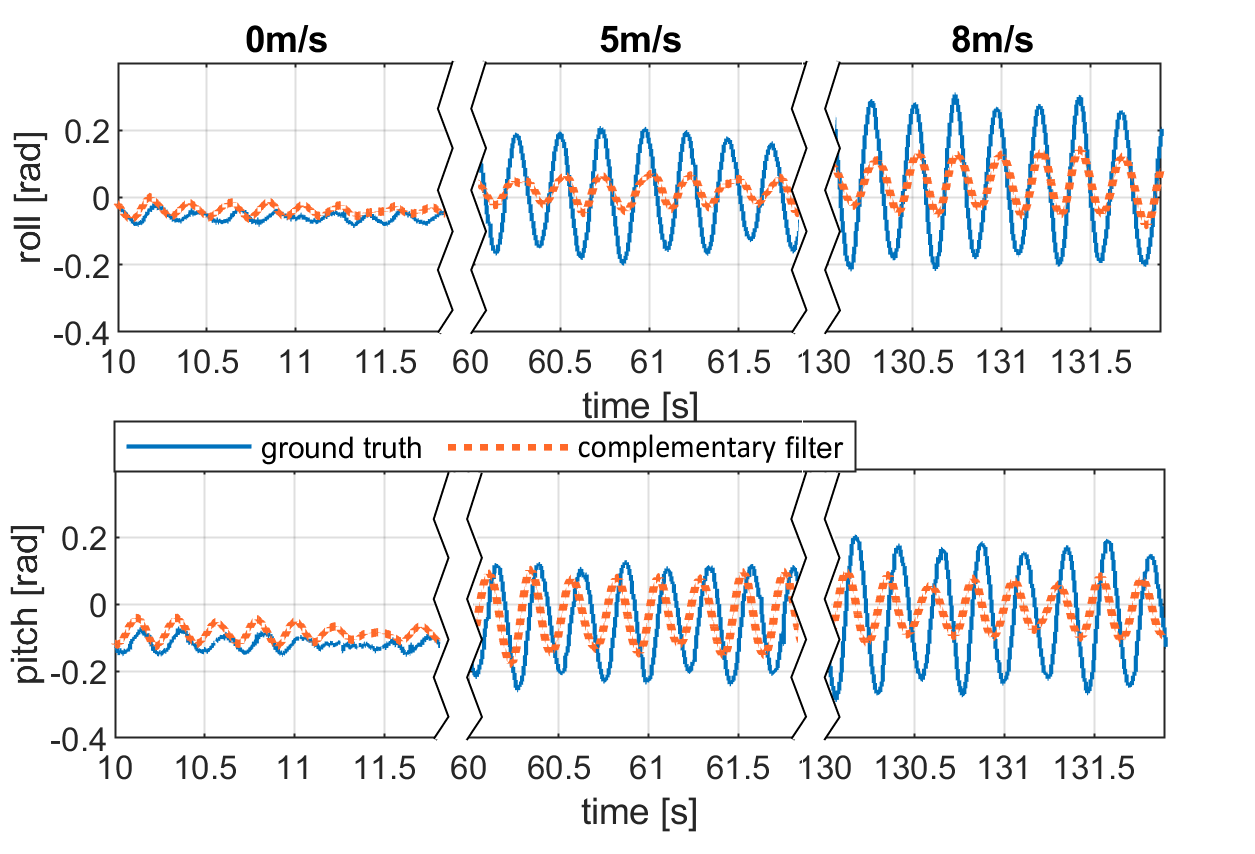}
    \caption{Comparison of pitch and roll estimations between the complementary filter and the ground truth in different flight speeds demonstrating degradation of the compensatory filter at these condition.}
    \label{fig:DRF_CF_ATT}
\end{figure}

\begin{figure}
    \centering
    \includegraphics[scale=0.84]{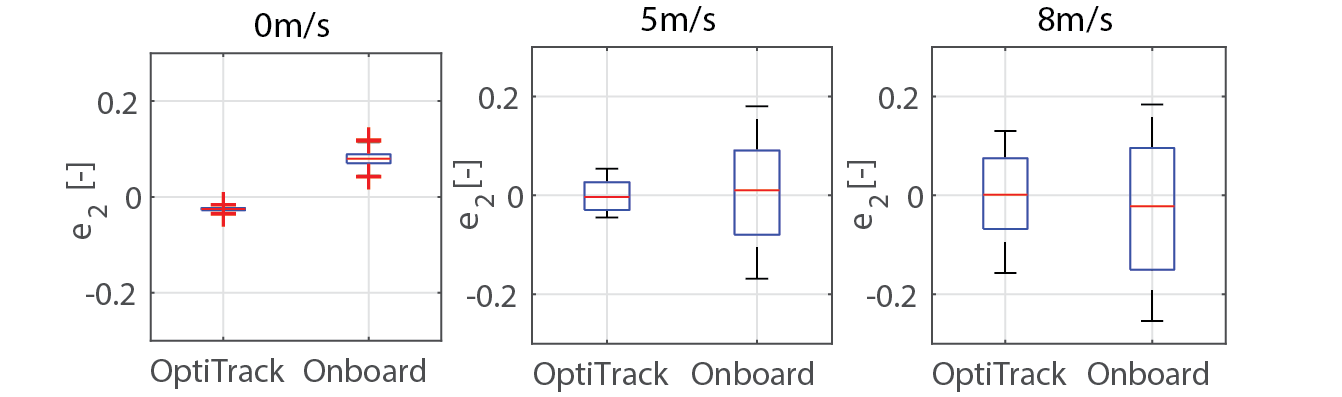}
    \caption{Boxplots comparing the tracking error of $y_2$, denoted by $e_2$, between flights with OptiTrack aided attitude estimation and those with onboard complementary filter, in different flight speeds.}
    \label{fig:y2error}
\end{figure}

In this setting, the INDI controlled drone achieved controlled flight at 8.4~m/s inside the wind tunnel, indicating robustness to significant attitude estimation errors. These errors can be seen in Fig.~\ref{fig:DRF_CF_ATT}, which compares the pitch and roll angles from the onboard complementary filter with the ground truth obtained with the OptiTrack system at three different flight speeds. As a consequence, the tracking errors of the reduced attitude were greater using the onboard complementary filter, especially at 0m/s and 5m/s as Fig.~\ref{fig:y2error} shows. Note that the increase of tracking error was less apparent at 8m/s where the drone was near the flight envelope boundary, because the controller performance was not only degraded by imperfect state estimations, but also motor limitations. 

We hypothesize that the degradation of the complementary filter is caused by the strong aerodynamic forces and centrifugal forces measured by the accelerometers. Improving the accuracy of the state estimator at high flight speeds and high angular rate conditions is out of the scope of this paper, but it is a highly recommended future research.

\section{CONCLUSION}
\label{sec:conclusion}

In this research, an incremental nonlinear fault-tolerant control method is developed for a quadrotor subjected to complete loss of two opposing rotors. The internal dynamics of this under-actuated control problem is analyzed. Thereby a criterion is given to select proper control outputs that guarantee the stability of internal dynamics, which has been validated in both simulations and flight tests. 
The control scheme can be generalized to a nominal quadrotor, or one with the loss of a single rotor.

The proposed method uses the incremental nonlinear dynamic inversion (INDI) approach to control the selected outputs. The INDI approach replaces non-input related model terms with sensor measurements, which reduces the model dependencies and consequently increases the robustness against wind disturbances in the high-speed flight regime.
Flight tests of a quadrotor with complete loss of two opposing rotors are conducted in an open jet wind tunnel. In the presence of significant aerodynamic effects, the control method is able to stabilize the quadrotor at over 8.0~m/s. Compared with the linear quadratic regulator (LQR) approach, the proposed method was found to have higher robustness against model uncertainties brought by the significant aerodynamic effects. 

Flights with imperfect state estimations from onboard sensors have been conducted. Flight data in the high-speed regime with onboard sensors reveal the adverse effects of aerodynamics on the state estimation. Future work is recommended to focus on improving the attitude estimation using onboard sensors by taking into account the effect of aerodynamics and high angular rate motion on the state estimator.

\addtolength{\textheight}{-0cm}   




\section*{ACKNOWLEDGMENT}


The authors would like to appreciate the support from the MAVLab and the help of these individuals: Leon Sijbers, Bram Strack van Schijndel, Bo Sun, Lan Yang and Prashant Solanki. We also appreciate the Aerodynamics Group of TU Delft for their permission to use the Open Jet Facility.

\bibliographystyle{ieeetr}
\bibliography{Reference.bib}

\begin{IEEEbiography}[{\includegraphics[width=1in,height=1.25in,clip,keepaspectratio]{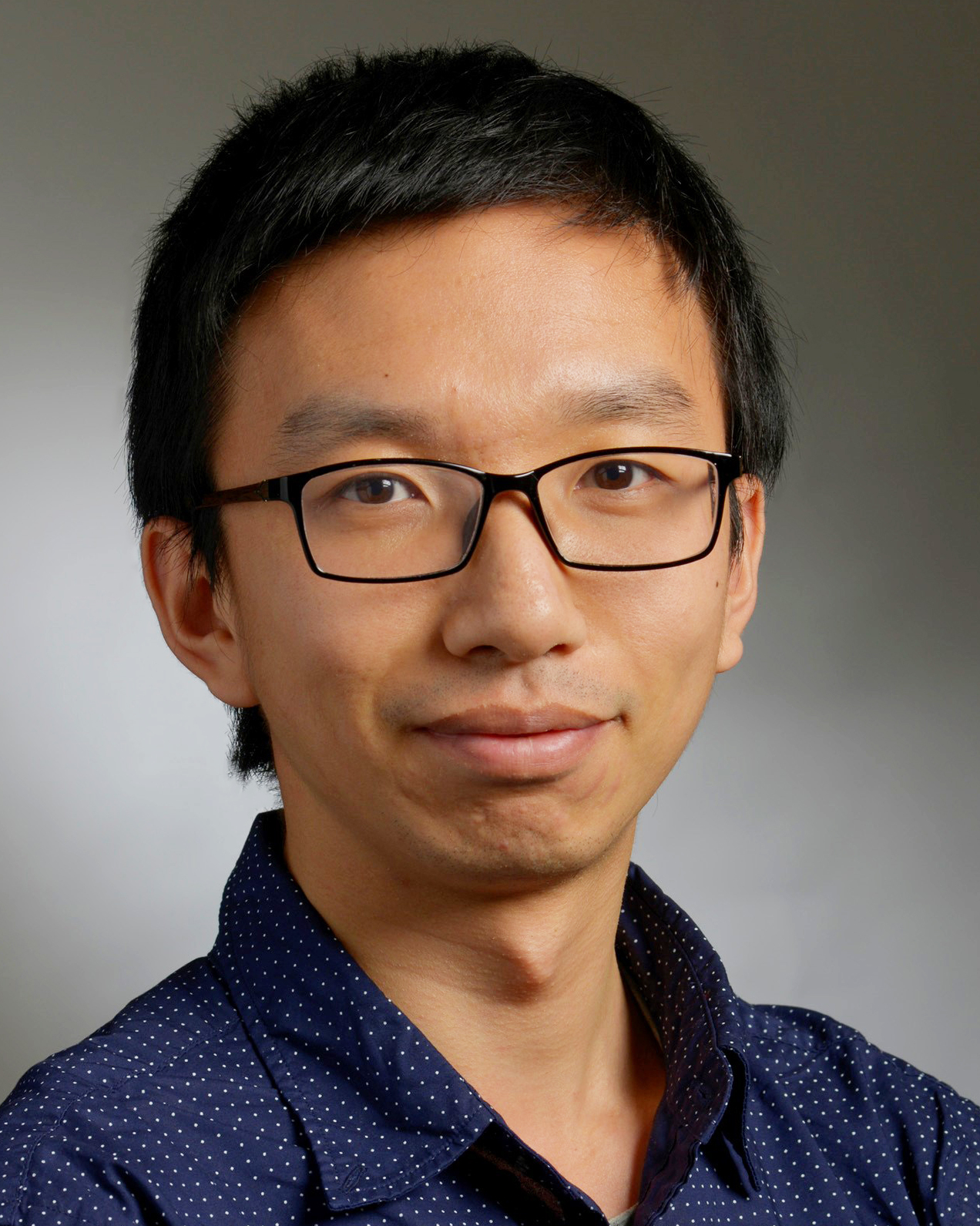}}]%
{Sihao Sun}
received his B.Sc and M.Sc degree in Aerospace Engineering from Beihang University, China, in July 2014 and February 2017 respectively. Since September 2016, he has been working towards the Ph.D. degree of Aerospace Engineering at Section of Control \& Simulation, Delft University of Technology, the Netherlands. In 2020, he was a visiting scholar at the Robotics and Perception Group, University of Zurich. His research interest includes system identification, aerial robotics, and nonlinear control.
\end{IEEEbiography}
\vskip 0pt plus -1fil
\begin{IEEEbiography}[{\includegraphics[width=1in,height=1.25in,clip,keepaspectratio]{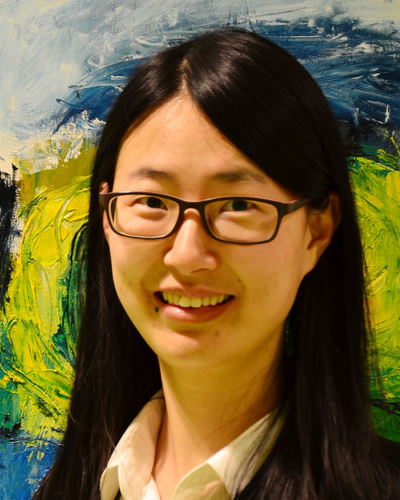}}]%
{Xuerui Wang}
received her Ph.D. degree in Aerospace Engineering from Delft University of Technology, the Netherlands, in July 2019. From May 2019 to May 2020, she was a Postdoctoral researcher with the Smart and Aeroelastic Structures Laboratory, Delft University
 of Technology, the Netherlands. Since May 2020, she has been with the faculty of Aerospace Engineering, Delft University of Technology, the Netherlands, as an Assistant Professor. Her research interests include nonlinear control, fault-tolerant control and aeroservoelasticity.
\end{IEEEbiography}
\vskip 0pt plus -1fil
\begin{IEEEbiography}[{\includegraphics[width=1in,height=1.25in,clip,keepaspectratio]{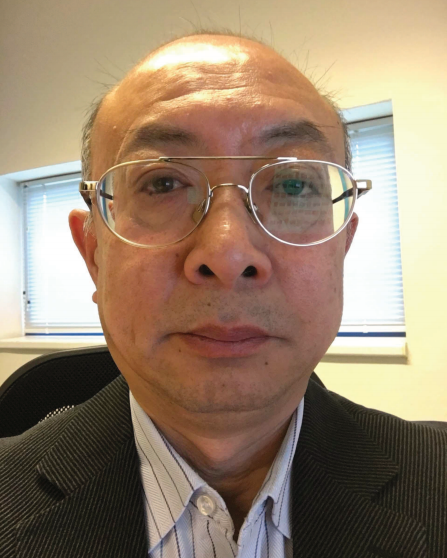}}]%
{Qiping Chu}
received his Ph.D degree from the
Faculty of Aerospace Engineering, Delft University of Technology in
1987. He is presently an associate professor and
the head of Aerospace Guidance, Navigation and
Control Cluster at the Section of Control and Simulation, Faculty of Aerospace Engineering, TU Delft.
The research interests of Dr. Chu are nonlinear
control, adaptive control, system identification and
state estimation with applications to fault tolerant
control, sensor integration, data analysis, and fault
detection/diagnosis.
\end{IEEEbiography}
\vskip 0pt plus -1fil
\begin{IEEEbiography}[{\includegraphics[width=1in,height=1.25in,clip,keepaspectratio]{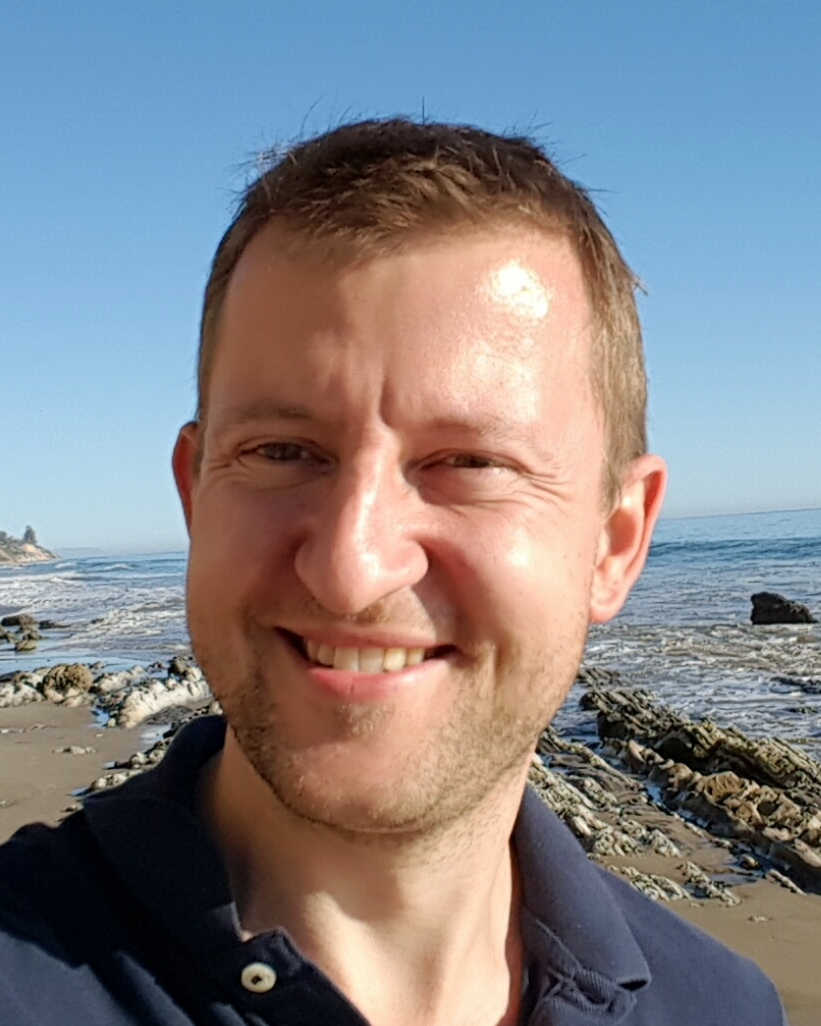}}]%
{Coen de Visser}
 received the M.Sc. degree from the Delft University of Technology in 2007. In 2011 he received his Ph.D degree from the faculty of Aerospace Engineering at the Delft University of Technology in The Netherlands. Between 2011 and 2012 he was a Postdoctoral Fellow at the Delft Center for Systems and Control where he worked on the development of a very large scale distributed control system for the adaptive optics system of the European Extremely Large Telescope. In 2015 and 2016 he was a visiting scholar at the University of California at Santa Barbara, and NASA Ames research centre respectively. Currently, he is an assistant professor at the Faculty of Aerospace Engineering at Delft University of Technology. His research interests are multivariate spline theory, aircraft system identification, flight envelope prediction and protection, and fault tolerant control. He has served in the international program committee of the European Guidance, Navigation, and Control conference between 2013 and 2019. He is the co-chair of the Council of European Aerospace Societies (CEAS) Technical Committee on Guidance Navigation \& Control, and serves in the American Institute of Aeronautics and Astronautics (AIAA) technical committee on Intelligent Systems. He is an Associate Editor for the International Journal of Micro Aerial Vehicles.
\end{IEEEbiography}
\end{document}